\documentclass[11pt]{article}
\usepackage[utf8]{inputenc}
\usepackage[a4paper,margin=1in]{geometry}
\usepackage{graphicx}
\usepackage{amsmath,amssymb}
\usepackage{booktabs}
\usepackage{caption}
\usepackage{subcaption}
\usepackage{float}
\usepackage{placeins}
\usepackage{hyperref}
\usepackage{setspace}

\newcommand{\panellabel}[1]{\makebox[\linewidth][l]{\textbf{#1}}\\[-0.35em]}
\newcommand{\figref}[2][]{\hyperref[fig:#2]{Fig.~#2#1}}
\newcommand{\tabref}[1]{\hyperref[tab:#1]{Table~#1}}

\title{Agentic Discovery of Cryomicroneedle Formulations}
\author{Hao Li, Lifu Du, Nurul Hameed, Shemonti Saha Authai,\\Zlata Stefanovic, Chenjie Xu\\[0.35em]\small Department of Biomedical Engineering, City University of Hong Kong\\\small 83 Tat Chee Avenue, Kowloon, Hong Kong SAR, China}
\date{}

\begin{document}
\maketitle

\begin{abstract}
Cryomicroneedles offer a route to minimally invasive intradermal delivery of living cells, but their cryogenic formulations must reconcile cell protection with constraints on toxicity and device fabrication. Here we report an AI-assisted, closed-loop workflow for cryomicroneedle cryoprotectant discovery that combines literature curation, Gaussian-process surrogate modelling, Bayesian optimization, and sequential wet-lab validation. A curated dataset of 198 mesenchymal stem-cell cryopreservation formulations from 42 studies was converted into 21 ingredient features and used to train an uncertainty-aware literature prior. This model captured moderate structure in the literature data but failed prospectively, motivating iterative wet-lab correction. Across ten validation iterations and 106 wet-lab observations, the model progressively adapted to cryomicroneedle-specific outcomes: batch RMSE decreased from 41.21 to 6.86 percentage points, later-stage rank correlations became consistently positive, and the cumulative wet-lab predicted-versus-measured summary reached $R^2 = 0.942$. The best validated formulation achieved 95.15\% post-thaw viability with low DMSO, ectoin, ethylene glycol, and fetal bovine serum. However, high viability alone did not ensure intact cryomicroneedle formation, highlighting the need for future multi-objective optimization. These results demonstrate that agent-assisted computational infrastructure can make data-efficient formulation discovery more accessible to labs with minimal data expertise in-house. Project code is available at \url{https://github.com/baitmeister/ML-for-CryoMN}.
\end{abstract}


\section{Introduction}

Cryomicroneedle is an emerging platform for the intradermal delivery of living cells. In this approach, a cryogenic formulation containing living cells is molded into microneedle templates and frozen, embedding viable cells directly within the needle tips. These frozen microneedle patches can either be prepared in advance and shipped at low temperature to the end-user to be later applied with minimal preparation or be directly fabricated by the end-user on the spot. Thus, cryomicroneedle combines the minimally invasive advantages of microneedle delivery with the logistical benefits of living cell cryopreservation.\cite{chang_cryomicroneedles_2021,zheng_insitu_cryomicroneedles_2024,bhatnagar_microneedles_2017}

A defining challenge in cryomicroneedle technology is the design of the cryogenic formulation. Unlike conventional cryopreservation media, which are optimized primarily to preserve cells during freezing and thawing, cryomicroneedle formulations must provide not only effective cryoprotection to maintain cell viability but also a mechanically stable structure capable of penetrating skin and dissolving rapidly after insertion. The combined biological and mechanical constraints increase the complexity of formulation design and create a large and heterogeneous formulation design space. Numerous candidate ingredients have been reported in the cryobiology literature, including permeating cryoprotectants such as dimethyl sulfoxide (DMSO) and ethylene glycol; membrane-stabilizing sugars like sucrose; osmoprotective solutes such as ectoine; and complex additives like serum. However, the performance of a formulation depends on nonlinear interactions among multiple components. Using exhaustive experiments to explore this combinatorial space is impractical, and intuition-driven approaches tend to prioritize familiar formulations that may overlook uncertain yet promising regions of the design space. As a result, identifying effective formulations for cryomicroneedle becomes difficult through conventional experimental design.\cite{whaley_overview_2021,murray_chemical_2022}

Machine learning offers promising strategies for navigating such complex formulation spaces by guiding the experimental exploration more efficiently. Instead of relying on exhaustive screening or empirically guided design, data-driven predictive models can learn relationships between formulation composition and experimental outcomes, allowing researchers to prioritize experiments that most likely improve performance or reveal new information.\cite{shahriari_human_2016,frazier_tutorial_2018,tom_self_driving_2024} However, implementing such data-driven approaches in practice presents a substantial barrier. Constructing a usable computational pipeline typically requires integrating literature data extraction, dataset curation, model training, uncertainty-aware experiment selection, iterative experimental updating, and analysis tools. Many experimental laboratories recognize the potential value of machine learning for experimental optimization but lack the computational expertise or engineering time needed to build and maintain this infrastructure. Thankfully, recent democratization of artificial intelligence (AI) suggests a possible solution to this barrier. AI agents are increasingly capable of interacting with software tools, generating code, and coordinating multi-step computational workflows under human supervision. Without replacing human researchers in scientific reasoning and higher-level decision-making, such systems function as programmable collaborators that provide additional guidance and help researchers construct the computational infrastructure required for modern data-driven discovery.\cite{wang_codeact_2024,yang_swe_agent_2024,lu_end_to_end_2026}

\begin{figure}[H]
\centering
\phantomsection\label{fig:1}
\includegraphics[width=0.97\linewidth]{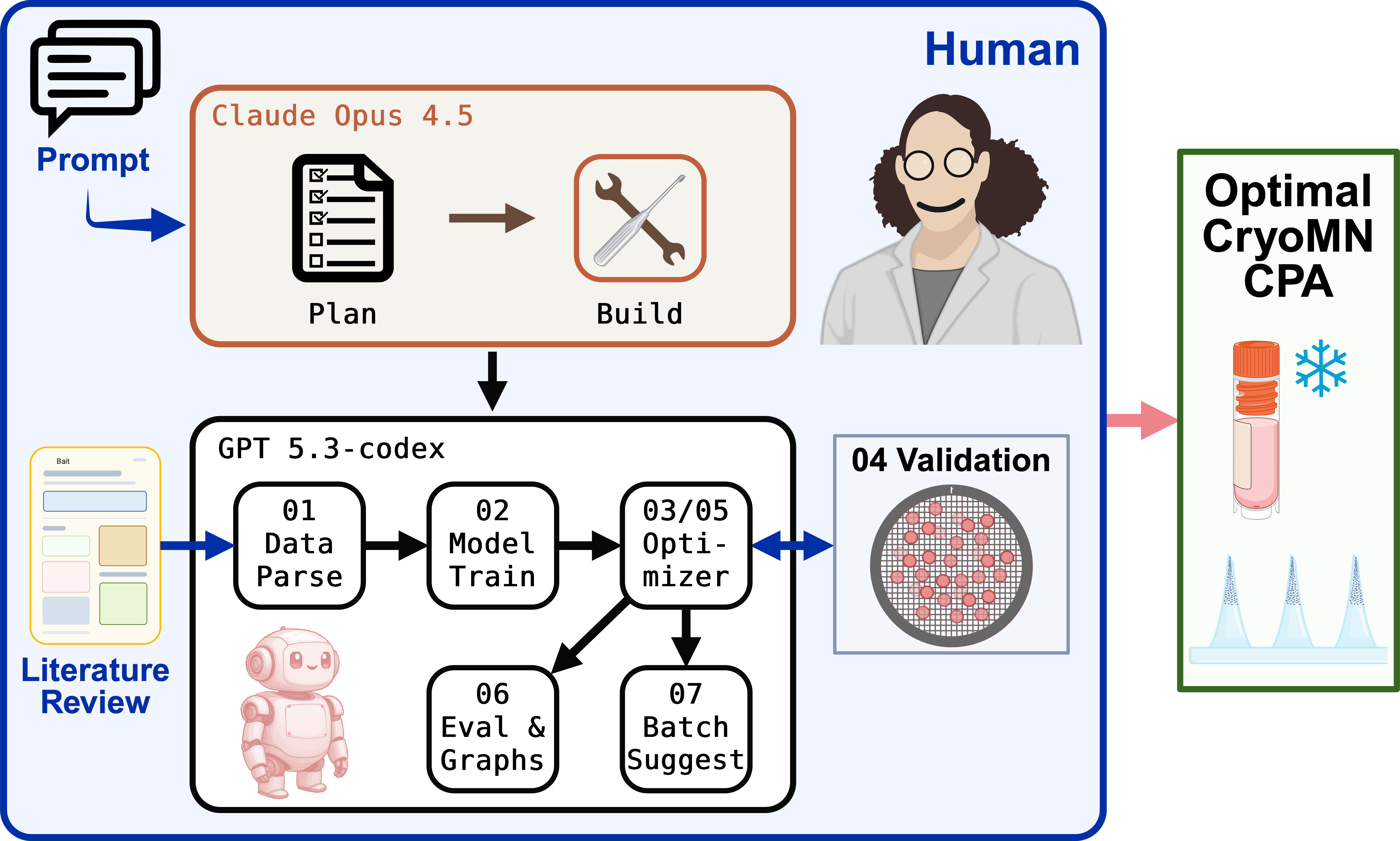}
\caption*{\textbf{Fig. 1 | }Schematic of the AI-assisted cryoprotectant formulation discovery workflow. Literature-derived formulation data are parsed into structured features, used to train Gaussian-process models, and iteratively updated with wet-lab validation results before new formulation batches are suggested.}
\end{figure}

Here we present a proof-of-concept study demonstrating how an AI-assisted computational workflow can support the discovery of cryomicroneedle formulations (\figref{1}). We develop an iterative machine-learning framework that integrates literature-derived data, probabilistic surrogate modeling, Bayesian optimization, and sequential wet-lab validation. Much of the supporting computational infrastructure is implemented through AI-assisted coding under human oversight, thereby significantly lowering the practical barrier to deploying such methods in experimental laboratories. Using this framework, we demonstrate the discovery of cryomicroneedle cryoprotectant formulations that reduce reliance on DMSO while maintaining high post-thaw cell viability. Other important attributes, including microneedle mechanical performance, remain crucial but are reserved for future multi-objective studies. Therefore, the aim of the present work is not to claim a final optimal cryomicroneedle formulation, but to illustrate how AI-assisted workflow construction can enable data-efficient experimental optimization in research environments where specialized computational expertise is limited.

\FloatBarrier

\section{Methods}

\subsection{AI-assisted workflow construction}

The computational pipeline was developed using AI-assisted coding under direct human supervision. In the initial one-shot prompt (Supplementary Methods), the scientific task was defined as a low-data formulation-optimization problem with four constraints: to maximize post-thaw viability, to minimize DMSO usage, to restrict candidate formulations to no more than 10 active ingredients, and to enable iterative updating after each wet-lab round. The agent, powered by Claude Opus 4.5 in Antigravity, proposed the plan to implement the overall workflow structure, including data parsing, surrogate modeling, optimization methods, candidate generation, model updating, and evaluation. After human revision of said plan, the agent proceeded to generate code, repository structure, and documentation. The generated materials from this one-shot prompt were reviewed and executed by the authors using Codex on GPT-5.3-codex. Additional features in the final program that did not appear in the initial repository, including evaluation and explainability, stage-indexed coefficient-of-determination $R^2$ plots, and next-batch formulation design, were prompted by the authors, coded, and implemented using Claude Opus 4.5 in Antigravity and GPT-5.3-codex in Codex.

\subsection{Literature dataset preparation}

The initial dataset was assembled from 75 primary research articles reporting a total of 499 cryoprotective formulations and viability outcomes (Supplementary Methods; Supplementary Data). We focused on the cryopreservation of mesenchymal stem cells (MSCs) because it is one of the model cell types in cryopreservation research. The literature review was done manually using keywords related to MSCs, including naming variants such as mesenchymal stromal cells and source-specific variants such as umbilical cord-derived MSCs or adipose-derived stem/stromal cells (ASCs), together with viability. Reported cryoprotective formulations and immediate post-thaw viability values were manually extracted as free text. Types of viability assays and cooling protocols were not modeled separately in the present proof-of-concept study. Such free-text formulation and viability descriptions were then parsed into a structured matrix in which each canonical ingredient occupied one feature column and each row represented one unique formulation. Synonymous ingredient names were merged into shared canonical identities, whereas basal media and buffer systems were excluded as formulation variables. Concentrations were harmonized into a mixed unit representation: ingredients with defined molecular properties were converted to molar units, whereas complex mixtures and polymers that could not be meaningfully represented in molar form were retained as percentages. Molecular-weight-specific polyethylene glycol species were tracked separately because their cryoprotective behavior may differ across size classes. Any duplicate formulation with discordant reported viabilities, largely due to different cooling methods, was collapsed by averaging the viability values because cooling methods are out of scope. Any relative viability (i.e., Formulation B has an $x$\% viability when normalized to Formulation A's viability) that might result in viability values greater than 100\% was excluded from model training. To reduce feature space and instability from very sparse variables, ingredients that appeared in fewer than 3 formulations were deleted. The whole data-cleaning process yielded the final 198 unique literature formulations described by 21 canonical ingredient features.

\subsection{Gaussian-process surrogate model}

Formulation performance was modeled using Gaussian-process regression. This approach was chosen because it is suitable for sparse experimental datasets and provides both a predicted mean and a predictive uncertainty for each candidate formulation. The initial literature-only surrogate used a Mat\'{e}rn kernel and standardized ingredient features before fitting. Initial model performance was estimated using five-fold cross-validation, in which the literature dataset was split into five subsets and each subset was predicted once by a model trained on the remaining four subsets.

\subsection{Candidate generation}

The trained Gaussian-process surrogate was used to propose candidate cryoprotective formulations for wet-lab testing. In the first five iterations, candidates were generated by random sampling of the feasible formulation space and then ranked by the Gaussian-process predicted mean viability. This strategy provided broad initial exploration when little cryomicroneedle-specific wet-lab feedback was available, allowing the model to learn the assay-specific landscape before being trusted to optimize it. In later iterations, candidate generation was replaced by Bayesian optimization. In this setting, the same Gaussian-process model served as the surrogate model, and an upper-confidence-bound (UCB) acquisition function detailed in Equation~\ref{eq:ucb} was used to combine predicted viability and uncertainty into a single search objective. In Equation~\ref{eq:ucb}, $x$ denotes a candidate formulation vector in feature space, $\mu{}\left(x\right)$ is the model-predicted mean viability for that candidate, $\sigma{}\left(x\right)$ is the model-predicted standard deviation for that candidate, and $\kappa{}$ is the exploration weight controlling how strongly uncertainty is rewarded relative to predicted mean performance, which we set to $\kappa{}=0.5$. The UCB score was computed in post-thaw viability percentage-point units. Model predictions were obtained as the sum of the literature-derived prior mean and the wet-lab residual-correction model, followed by the stage-specific additive mean correction and multiplicative uncertainty correction stored in the model metadata. Thus, both $\mu(x)$ and $\sigma(x)$ in Equation~\ref{eq:ucb} are reported on the calibrated viability-percent scale.

\begin{equation}
a_{\mathrm{UCB}}\left(x\right)= \mu{}\left(x\right)+\kappa{}\cdot{}\sigma{}\left(x\right)\label{eq:ucb}
\end{equation}

Then, differential evolution was used as the numerical optimizer to search for formulations with high acquisition scores under formulation constraints. Candidate formulations were required to satisfy practical constraints, including a maximum of 10 active ingredients and an upper DMSO bound. We set up two different DMSO upper limits, thus generating two different pools of candidate files: "general" at 5\% DMSO and "DMSO-free" at 0.5\% DMSO\@. More detailed reasoning on optimization methods, search bounds, acquisition parameters, and batch-selection rules are provided in Supplementary Methods.

After the formulation candidates were generated, the final wet-lab slate was selected by a separate formulation recommendation step. In the first five iterations, where random search was used, the selection was done manually by picking candidates with high predicted viability yet not overly similar to other formulations. In later iterations, this step involved combining high-performing exploitation candidates with exploration candidates designed to expose model blind spots. Exploitation candidates were selected from the highest-ranked Bayesian-optimization outputs after removing already tested formulations, whereas exploration candidates were generated from residual patterns observed in previous wet-lab stages. Formulations that had performed better than predicted were used as local anchors, and historically underpredicted regions of formulation space were used to define blind-spot probes. Additional coverage probes were included to expand support in sparse regions of the observed formulation space. In this way, the formulation recommendation step translated model predictions and model failures into a balanced set of wet-lab validation candidates, intended to both identify strong formulations and improve future model performance.

\subsection{Wet-lab validation}

Selected formulations per recommendation were prepared. Umbilical cord-derived human mesenchymal stem cells at either the 7\textsuperscript{th} or 8\textsuperscript{th} passage were used to conduct the viability validation. These cells were cultured in low-glucose Dulbecco's modified Eagle medium (DMEM) with GlutaMAX and pyruvate (Gibco, Cat. 10567014) until 80\% confluence before being trypsinized for viability assay. Polydimethylsiloxane (PDMS) negative molds used to fabricate cryomicroneedles were affixed inside the wells in 6-well plates. Detailed specifications of the PDMS mold and the resulting cryomicroneedles can be found in Supplementary Methods. A 700 \ensuremath{\mu}L aliquot of the cell suspension solution prepared using the selected formulation at around $1\times{}10^{6}$ cells/mL was added to PDMS negative molds, and the plates containing the molds were then centrifuged once for 3 minutes at 400 $g$. Afterwards, no air bubbles or hollow cavities near the needle channels were visible. Excess liquid outside of the needle tips was wicked away with tissue paper. Needle tips were clearly filled with cell-containing candidate cryoprotectants. Another 700 \ensuremath{\mu}L of the respective cell-containing cryoprotectant solution was added to each mold. The 6-well plates containing molds were transferred to a 4$^\circ$C refrigerator for 30 minutes, then to a -20$^\circ$C freezer for 2.5 hours, before being transferred to a -80$^\circ$C freezer for another 2.5 hours. Afterwards, these cryomicroneedle-containing plates were directly transferred to a 37$^\circ$C floating water bath within 15 seconds, where the cryomicroneedles were thawed inside the wells of the 6-well plates. Thawed solutions were transferred to centrifuge tubes for centrifugation at 400 $g$ for 4 minutes. The supernatant was then removed, and the cells were resuspended with 2 mL staining cocktail in a 12-well plate. This cocktail was prepared to have 1 \ensuremath{\mu}M CellTracker Green CMFDA and 1 drop of NucBlue Live ReadyProbes reagent containing Hoechst 33342 per mL in phosphate-buffered saline (PBS). Immediately after, the 12-well plates were transferred to the incubator (37$^\circ$C, 5\% CO\textsubscript{2}, 95\% relative humidity) for 45 minutes for staining. Afterwards, 18 \ensuremath{\mu}L of thawed cryomicroneedle solution was sampled from each well into a Countess Cell Counting Chamber Slide for viability assay. Post-thaw viability was defined as the number of CMFDA-positive (alive) cells divided by the number of Hoechst-positive (all) cells in the Countess image field. For each formulation, five image replicates were analyzed using ImageJ, and the average was used as the viability for that formulation.

\subsection{Feedback and model update}

Measured post-thaw viability values were incorporated into the next model update. The literature-only model generated the first validation batch, which is referred to as stage 0 or iteration 0. After this stage, the model was updated using a prior-mean correction strategy: the literature-trained Gaussian process was retained as the baseline predictor, and a second Gaussian process was fitted to the residuals between measured wet-lab viability and literature-model prediction. The final prediction was the sum of the literature-derived prediction and the wet-lab residual correction. After the model update, iteration 1 began. Later iterations followed the same closed-loop structure: update the surrogate, generate candidates, test selected formulations, and update the model again using wet-lab results.

This update strategy allowed the model to preserve broad formulation trends learned from the literature while adapting to cryomicroneedle-specific wet-lab outcomes. Wet-lab observations were given a default 50-fold greater local influence than literature data through source-specific noise assumptions. Wet-lab generalization was estimated by cross-validating with only the wet-lab rows while retaining all literature rows in the training set for every fold, and the residual diagnostics from this process were then converted into two post-hoc calibration terms applied downstream to all reported predictions: an additive mean correction and a multiplicative uncertainty correction. Details of the implementation, other parameters, and the cross-validation protocol can be found in Supplementary Methods.

\subsection{Evaluation and explainability}

After the wet-lab viability feedback, model evaluation was performed at the stage level using frozen checkpoints. Each saved model was evaluated against the wet-lab batch that had been selected from that stage, preserving the historical decision context. For example, the literature-only model was evaluated against the first wet-lab batch chosen from the literature-only candidate files (stage 0), whereas the frozen iteration 3 model was evaluated against the wet-lab formulations that were selected from the iteration 3 candidate outputs, not against formulations proposed later by iteration 4 or iteration 5 models. This prevents later model revisions from rewriting the evidence for earlier decisions and ensures that each stage is judged on the prospective experiment set it truly generated.

For each stage batch, we reported various metrics, including root mean squared error (RMSE), mean absolute error (MAE), ranking accuracy, and Spearman's and Kendall's rank correlation coefficients for that specific batch. Moreover, we reported the coefficient of determination using a prospective-cumulative $R^2$ protocol. At each stage $k$, predictions from the frozen stage model were evaluated against the cumulative wet-lab set available up to stage $k$. This design preserves temporal ordering while increasing effective sample size and outcome-range coverage, which makes the explained-variance estimate more stable and interpretable across iterations. We did not treat batch-specific $R^2$ as a primary endpoint because individual wet-lab batches were small, often with only a few formulations validated, and $R^2$ is highly sensitive to both outliers and within-batch response variance under those conditions. As a result, batch $R^2$ could fluctuate sharply, including sign reversals, without reflecting a meaningful change in practical predictive utility. In this setting, reporting only batch $R^2$ would therefore be noisy and potentially redundant relative to the broader prospective-cumulative trend.

To make the model's decisions and reasoning more interpretable, an explainability module was used to assess the active iteration-specific observed context. Graphs that showed feature importance, Shapley additive explanations (SHAP) summaries, support-aware marginal effects, pairwise interaction contours, acquisition landscapes, and uncertainty diagnostics were generated to identify dominant formulation variables, visualize where the model was interpolating versus extrapolating, and assess whether predictive uncertainty remained useful for experimental decision-making. Detailed definitions of the evaluation metrics, prospective-cumulative $R^2$ implementation, and explainability plots are provided in Supplementary Methods.

\section{Results}

The complete agent-assisted workflow connecting literature curation, model training, candidate generation, wet-lab validation, and batch recommendation is summarized in \figref{1}.

\subsection{Initial baseline architecture}

The one-shot prompt generated a baseline that was immediately organized as a modular closed-loop pipeline rather than as an isolated analysis notebook. The prompt defined the problem as a low-data cryoprotectant optimization, constrained candidate recipes to no more than 10 active ingredients, asked for DMSO minimization and viability maximization, and required the model to evolve after wet-lab validation. In response, the first substantive repository commit on January 24, 2026 added a four-stage architecture: free-text formulation parsing, Gaussian-process surrogate training, constrained candidate generation, and a validation/update loop. Each stage was placed in a separate source directory with a companion README, and the same commit added requirements, processed data, a trained baseline model, candidate tables, and a wet-lab validation template.

The data layer converted unstructured formulation descriptions into a machine-learning matrix by canonicalizing ingredient synonyms, excluding basal media and buffers, harmonizing concentrations where possible, retaining DMSO as both a feature and a design constraint, and averaging duplicate formulation entries with discordant viabilities. The modeling layer then fit a standardized Gaussian-process regressor with a Mat\'{e}rn kernel, which was selected because the literature dataset was small and because predictive uncertainty was needed for sequential experiment selection. The first baseline model tracked 21 ingredient features and already produced cross-validation diagnostics, feature-importance output, and serialized model artifacts.

The first optimization layer was already split into two candidate streams: a general low-DMSO pool with an upper DMSO bound and a near-DMSO-free pool. The first generated summaries ranked predicted formulations with uncertainty estimates and ingredient counts, while the validation module created the file interface needed to add measured post-thaw viability and retrain the model. The early commit history on GitHub showed the limitations of this baseline as well as its usefulness; yet within the next three days, follow-up commits corrected candidate formatting, added a dedicated differential-evolution Bayesian-optimization module, introduced explainability plots, and implemented a prior-mean wet-lab correction model. Therefore, the one-shot output did not produce the final scientific workflow, but it did generate the working scaffold from which the later iteration-aware, calibrated, and explainable system was built.

\vspace{1.2em}
\begin{table}[H]
\centering
\phantomsection\label{tab:1}
\caption*{\textbf{Table 1 | }Top 15 ingredients sorted by appearance in unique formulations reported in the literature.}
\begin{tabular}{@{}lrr@{}}
\toprule
\textbf{Ingredient} & \textbf{Appearances} & \textbf{Highest viability reported (\%)} \\
\midrule
Dimethyl sulfoxide (DMSO) & 89 & 100 \\
Fetal bovine serum (FBS) & 58 & 92 \\
Sucrose & 31 & 93.4 \\
Trehalose & 28 & 99 \\
Ethylene glycol & 20 & 96 \\
Human serum albumin (HSA) & 20 & 95.7 \\
Glycerol & 18 & 83 \\
Hydroxyethyl starch (HES) & 17 & 100 \\
Human serum & 13 & 95.7 \\
Polyethylene glycol (PEG; $\sim$3350 Da) & 13 & 94 \\
Polyvinylpyrrolidone (PVP) & 13 & 72.5 \\
Methylcellulose & 11 & 91 \\
Propylene glycol & 8 & 83.4 \\
Ectoin & 6 & 96 \\
Betaine & 5 & 84 \\
\bottomrule
\end{tabular}
\end{table}

\subsection{Dataset cleanup and literature formulation landscape}

Our dataset comprises 198 unique formulations from 42 independent studies retained after cleaning the original 393 formulations (Supplementary Data). The overall post-thaw viability across all formulations demonstrated significant variance, averaging 63.88\% (median: 70.00\%, range: 0.00--100.00\%). Analysis of the distribution of standard cryoprotective agents reveals DMSO as the most prominent component, present in 89 formulations (44.9\%). Other frequently utilized additives included fetal bovine serum (FBS) (29.3\%), sucrose (15.7\%), trehalose (14.1\%), ethylene glycol (10.1\%), and human serum albumin (HSA) (10.1\%). The dataset contains a slight majority of DMSO-free formulations (n = 109, 55.1\%). While popular, there remains an efficacy gap between the two categories: DMSO-containing formulations (n = 89) demonstrated a higher average post-thaw viability of 71.90\%, whereas DMSO-free formulations (n = 109) yielded a notably lower average post-thaw viability of 57.33\%. A summary of the literature data is presented in \tabref{1}.

A critical subset of 59 formulations (29.8\%) achieved excellent post-thaw cell viabilities of \ensuremath{\geq}80\%. Within this high-performing threshold, the reliance on DMSO was pronounced, occurring in 42 (71.2\%) of these top formulations. Other recurring components in this high-efficacy tier included FBS, trehalose, hydroxyethyl starch (HES), and ethylene glycol. These results established both the usefulness and the bias of the literature prior: the model could learn meaningful formulation structure, but a direct literature-trained optimum would likely over-favor DMSO-containing recipes.

\subsection{Initial model and first wet-lab validation}

To establish a pre-experimental baseline, we first trained a Gaussian process model using the literature-curated formulation dataset. On five-fold cross-validation, this model achieved an RMSE of 20.25 \ensuremath{\pm} 1.99 with a mean $R^2$ of 0.247 \ensuremath{\pm} 0.096, and when fit to the full literature set it reached a training RMSE of 14.55 and $R^2 = 0.618$ (\figref[a]{2}). These results indicated that the literature data contained recoverable predictive structure, but only with moderate fidelity. The literature-only model's SHAP importance showed that the initial model was strongly shaped by ingredients that were common or historically successful in the literature record (\figref[c]{2}).

\begin{figure}[!p]
\centering
\phantomsection\label{fig:3}
\includegraphics[height=0.33\textheight,keepaspectratio]{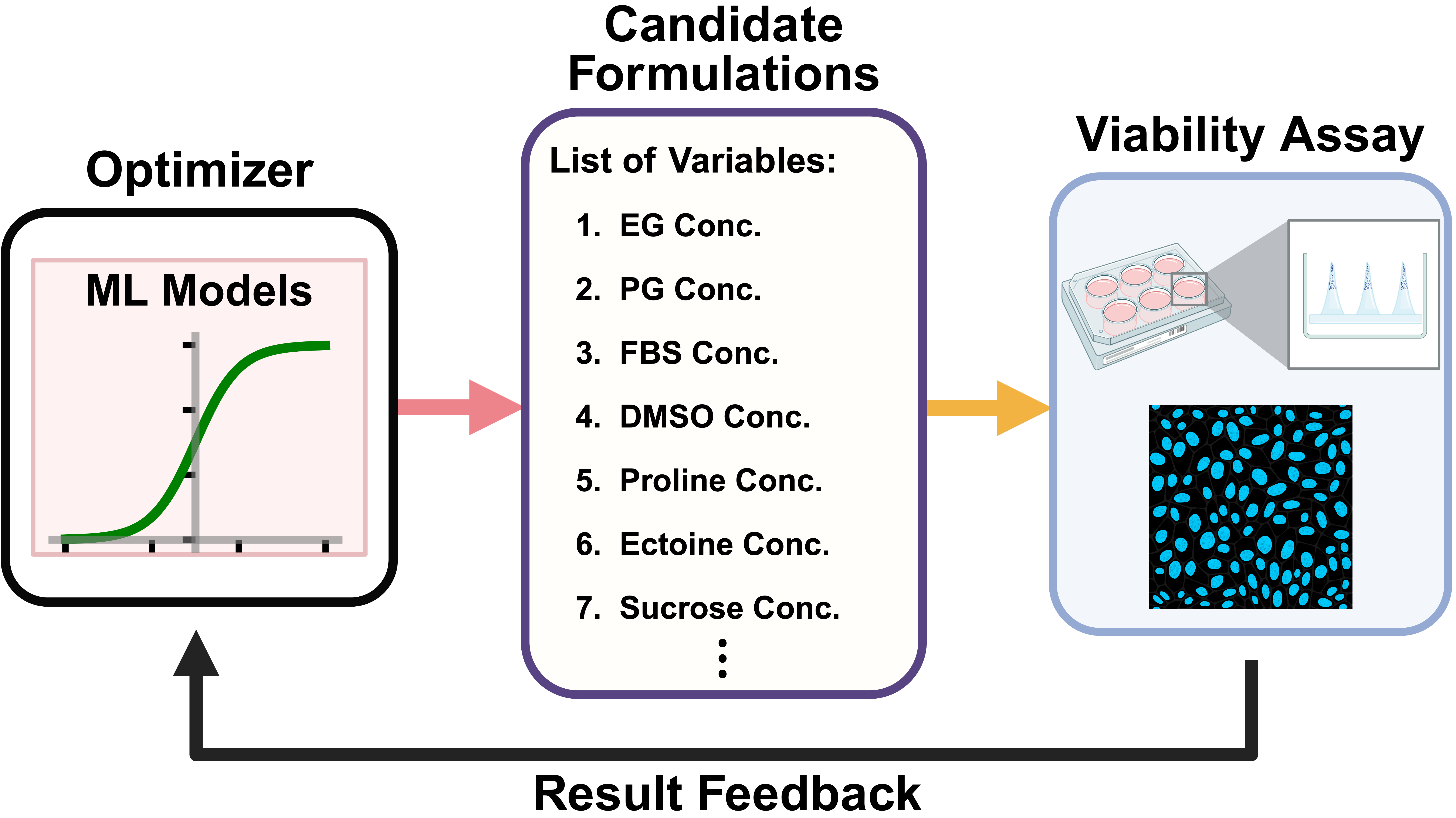}
\caption*{\textbf{Fig. 3 | }Schematic of the iterative optimization process. An optimizer based on the trained model generates candidate formulations for viability testing. The measured results are then used to update the trained models.}
\end{figure}

That performance did not translate prospectively. In the first wet-lab validation on the literature-only model (iteration 0) using the iterative process described in \figref{3}, the model's error increased sharply on 18 candidate formulations, with RMSE rising to 41.21 while $R^2$ fell to -2.52, and rank-based agreement was poor (\figref[b]{2}, \figref{4}). We attribute this drop primarily to the heterogeneity and reporting bias of literature-derived data. The aggregated studies span different mesenchymal stem cell sources, cryopreservation workflows, assay endpoints, post-thaw handling conditions, and reporting standards, meaning that nominally similar formulations may reflect substantially different experimental contexts. The model update after this first batch changed the attribution pattern, as shown by the post-update SHAP importance (\figref[d]{2}), supporting the need for iterative wet-lab feedback rather than one-shot literature extrapolation.

\makeatletter
\newlength{\savedfptop}
\newlength{\savedfpsep}
\newlength{\savedfpbot}
\setlength{\savedfptop}{\@fptop}
\setlength{\savedfpsep}{\@fpsep}
\setlength{\savedfpbot}{\@fpbot}
\setlength{\@fptop}{0pt}
\setlength{\@fpsep}{24pt}
\setlength{\@fpbot}{0pt plus 1fil}
\makeatother

\begin{figure}[!p]
\centering
\phantomsection\label{fig:2}
\begin{subfigure}{0.48\textwidth}
\centering
\panellabel{a}
\includegraphics[width=\linewidth]{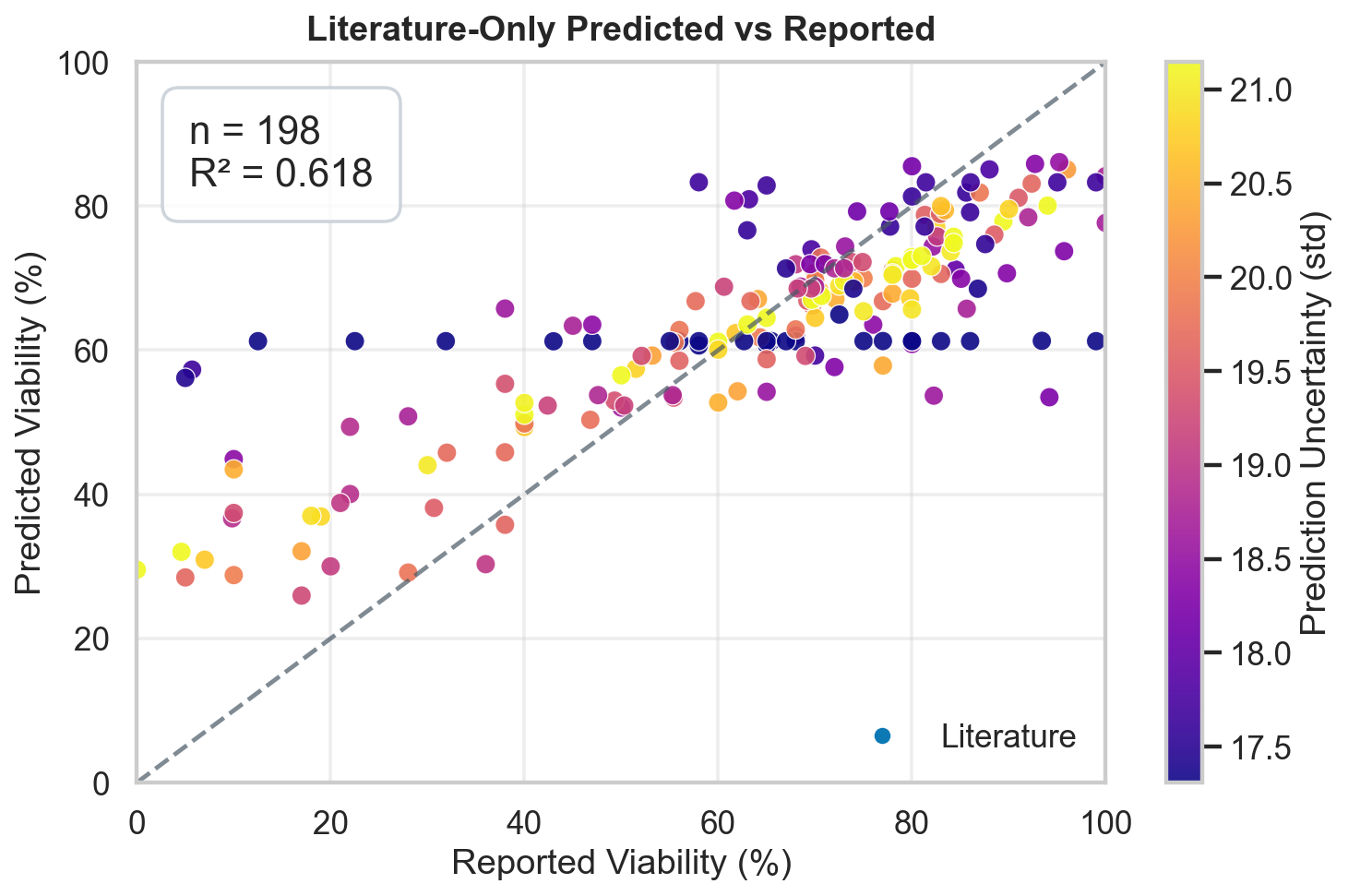}
\end{subfigure}
\hfill
\begin{subfigure}{0.48\textwidth}
\centering
\panellabel{b}
\includegraphics[width=\linewidth]{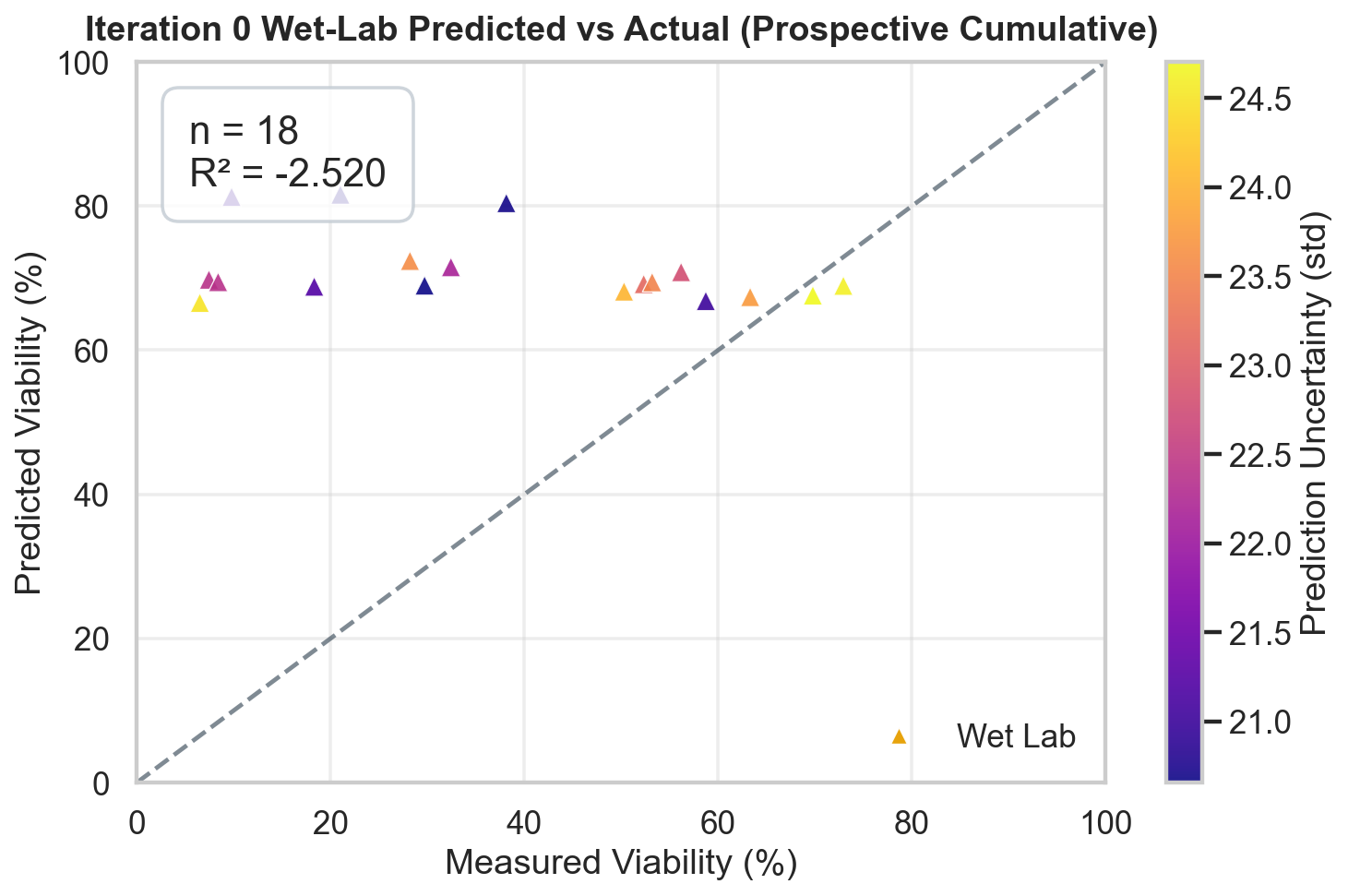}
\end{subfigure}

\begin{subfigure}{0.48\textwidth}
\centering
\panellabel{c}
\includegraphics[width=\linewidth]{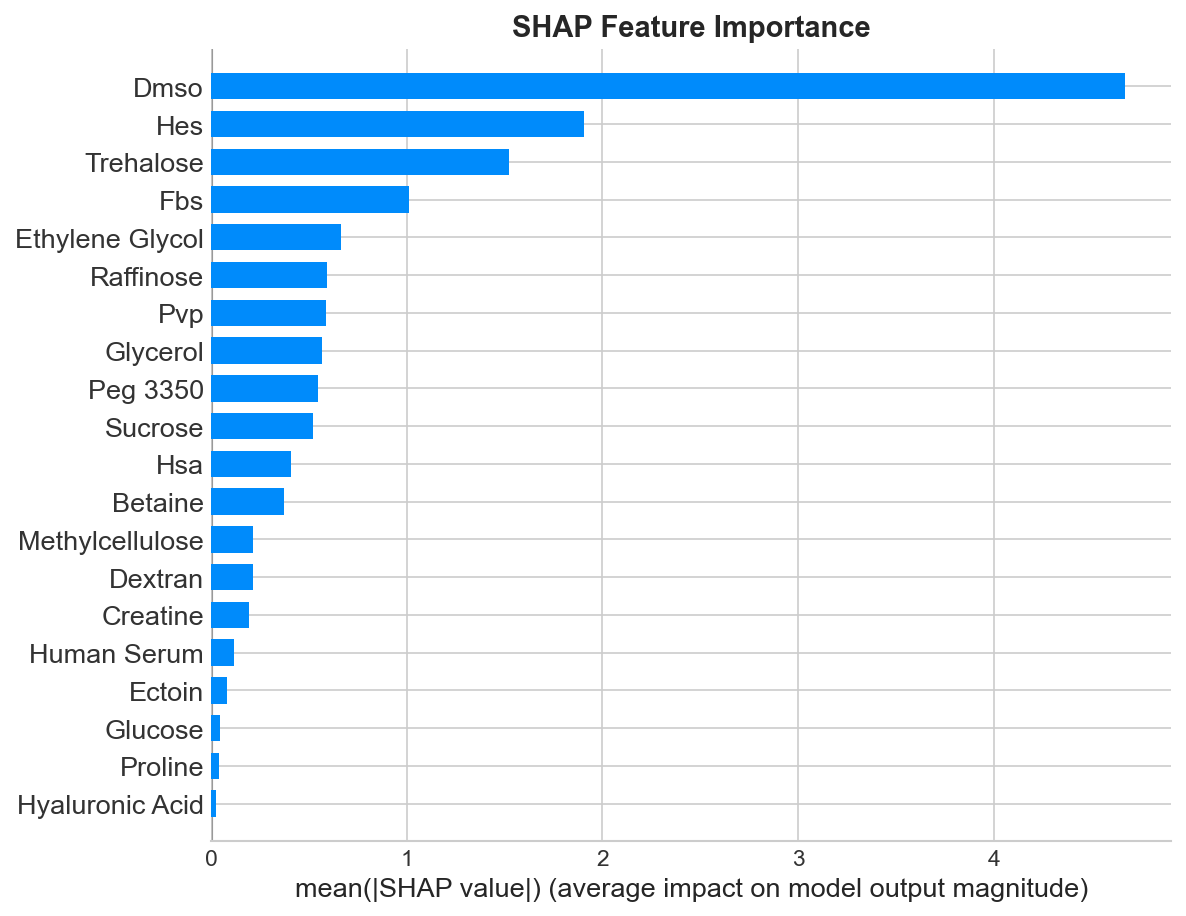}
\end{subfigure}
\hfill
\begin{subfigure}{0.48\textwidth}
\centering
\panellabel{d}
\includegraphics[width=\linewidth]{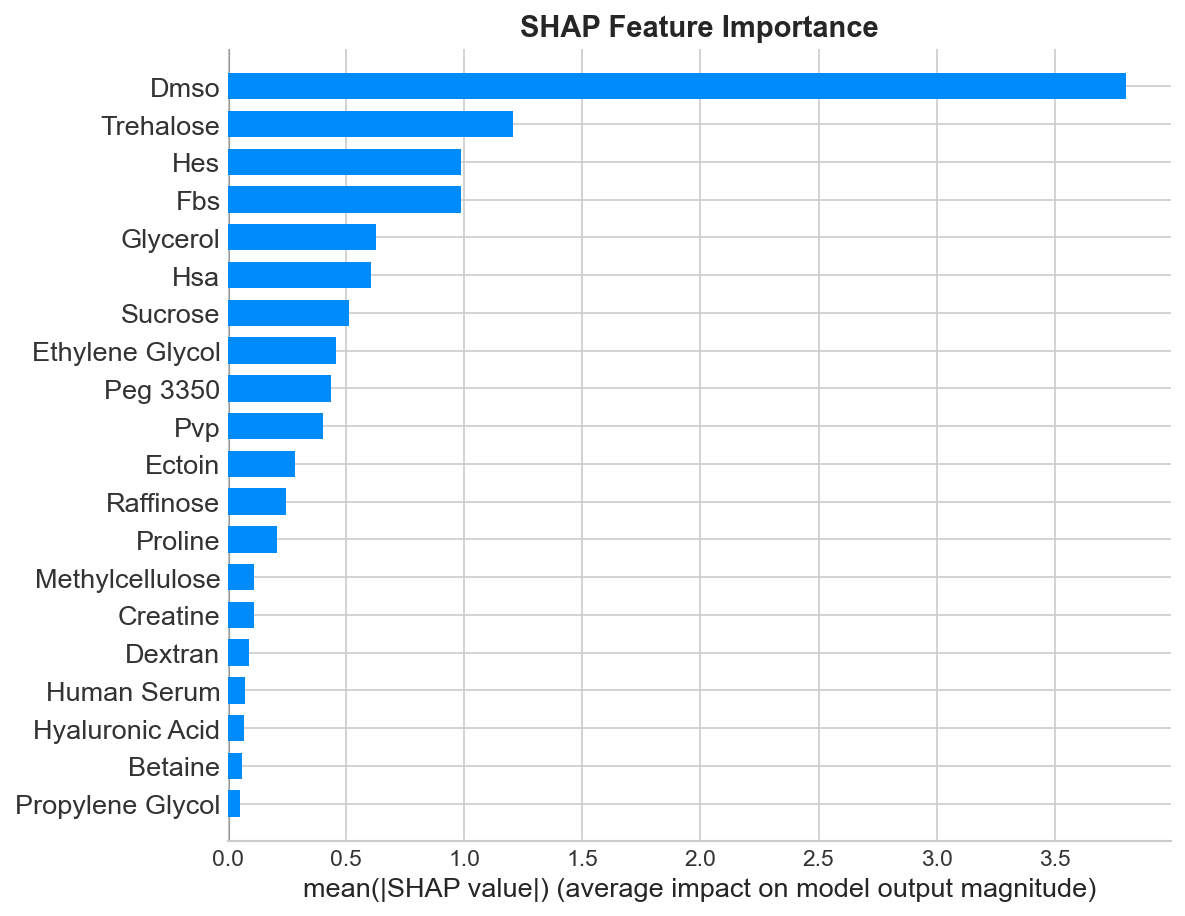}
\end{subfigure}
\caption*{\textbf{Fig. 2 | }Initial model and first wet-lab validation: a, literature-only predicted versus reported viability; b, wet-lab predicted versus measured viability for the first validation batch; c, SHAP importance for the literature-only model; d, SHAP importance after model update using the first wet-lab validation batch results. In a and b, each dot represents a formulation's viability.}
\end{figure}

\FloatBarrier
\makeatletter
\setlength{\@fptop}{\savedfptop}
\setlength{\@fpsep}{\savedfpsep}
\setlength{\@fpbot}{\savedfpbot}
\makeatother

\subsection{Model after ten iterations of wet-lab validation}

\begin{figure}[H]
\centering
\phantomsection\label{fig:4}
\includegraphics[width=\linewidth]{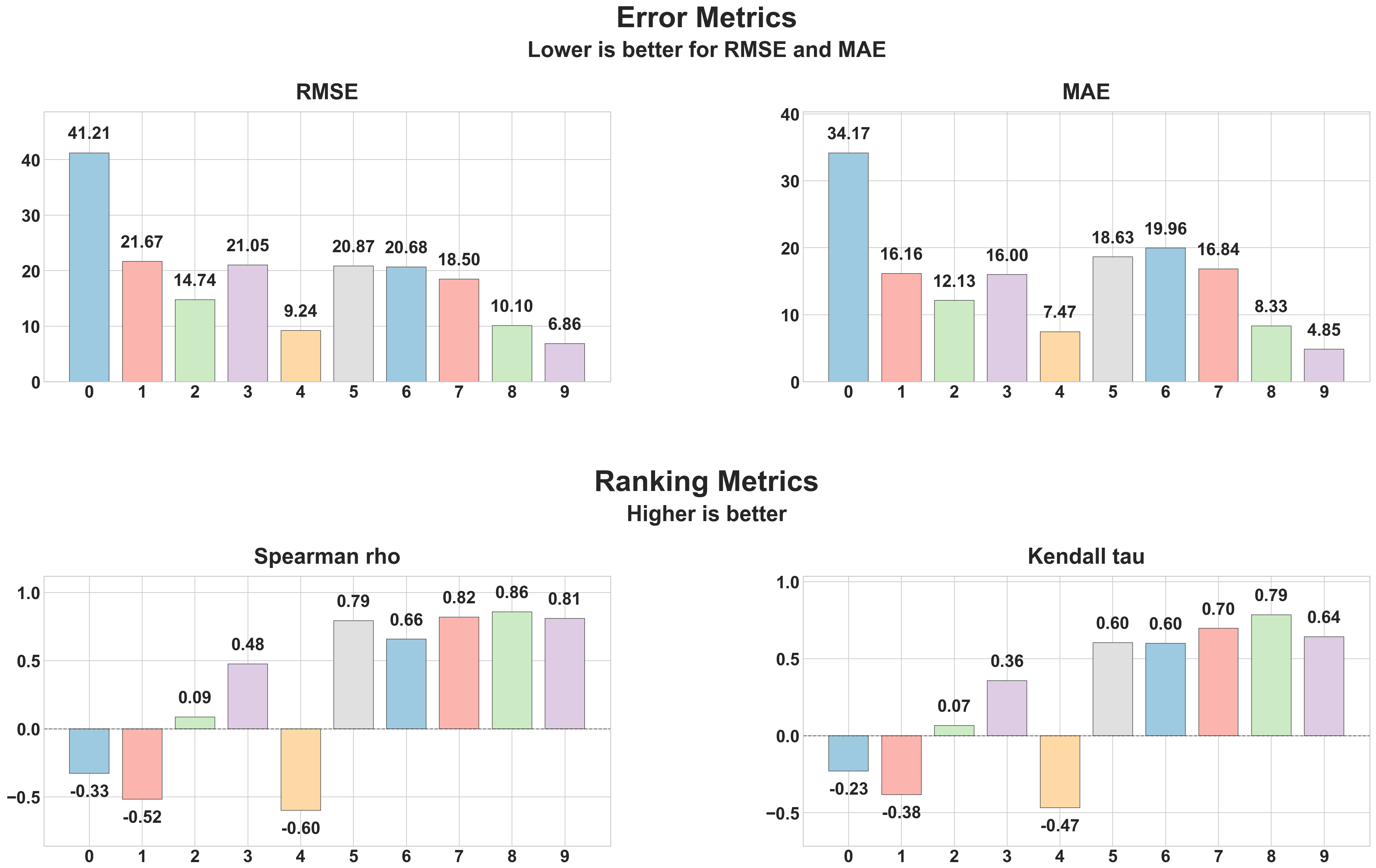}
\caption*{\textbf{Fig. 4 | }Batch-level model evaluation across validation stages. Frozen model checkpoints were evaluated against the wet-lab batches generated at the corresponding stage. RMSE and MAE are reported in viability percentage points; lower values indicate smaller prediction error. Spearman's $\rho$ and Kendall's $\tau$ measure rank agreement within each batch; higher values indicate better preservation of formulation ordering. Because individual wet-lab batches were small, these metrics should be interpreted as stage-level diagnostics rather than progressive performance curves.}
\end{figure}

Across the closed-loop validation campaign, the model was updated after each wet-lab batch and progressively shifted from a literature-only predictor toward a wet-lab-corrected surrogate. In total, the validation table contains 106 wet-lab observations after the literature-derived training set. The practical outcome improved relative to the literature-only batch: the best formulation in the literature-only batch reached 72.93\% viability, iteration 1 increased the best-so-far value to 79.09\%, and iteration 5 identified the highest observed formulation, EXP5110 with 21.0 mM DMSO, 291.1 mM ectoin, 1.79 M ethylene glycol, and 5.4\% FBS, achieving 95.15\% viability (Supplementary Data). We would like to note that iteration 5's elevated viability should be interpreted cautiously because it may reflect a combination of model-guided enrichment and batch-level experimental variation. The high-performing formulations clustered around a related low-DMSO ectoin/ethylene glycol/FBS region, but they were tested within a narrow wet-lab window in which subtle differences in cell condition, handling, and measurement could systematically shift viability. Thus, this result is useful evidence for a promising local formulation region, but should not be taken as a standalone demonstration of global model improvement. Over the course of the validation loop, model error also improved. Batch RMSE fell from 41.21 in the literature-only prospective test to 21.67 after iteration 1, 14.74 after iteration 2, 10.10 after iteration 8, and 6.86 after iteration 9, while later-stage rank agreement became consistently positive (\figref{4}).

A comprehensive look across all batches showed the same overall trend while retaining the expected noise of small experimental batches (\figref{4}). The earliest stages (iteration 0-4) using random search were unstable: models had erratic ranking behavior (Spearman's $\rho$ and Kendall's $\tau$); even at iteration 4, the model still had a negative rank correlation despite iteration 4's low RMSE due to the narrow viability range of formulations in that batch. In contrast, later Bayesian optimization stages (iteration 5-9) showed exploratory behavior that improved ranking significantly: Spearman's $\rho$ reached 0.79, 0.66, 0.82, 0.86, and 0.81 across iterations 5-9. Although the exploration formulations drove up error in the early Bayesian optimization stages, error eventually dropped with more iterations. These metrics show that the loop became practically better at understanding the viability range and at nominating testable formulations with good viability. In the end, iteration 9 had a batch RMSE of 6.86, and the cumulative wet-lab predicted-versus-measured summary containing 106 wet-lab observations reached an $R^2$ of 0.942 (\figref[a]{5}). After iteration 9's model was updated with wet-lab results, the iteration 10 SHAP feature importance was plotted to further demonstrate the shift from the initial literature model, with ethylene glycol and ectoin emerging as major contributors alongside the original DMSO-rich literature signal rather than remaining subordinate to it (\figref[b]{5}).

\begin{figure}[H]
\centering
\phantomsection\label{fig:5}
\begin{subfigure}{0.48\textwidth}
\centering
\panellabel{a}
\includegraphics[width=\linewidth]{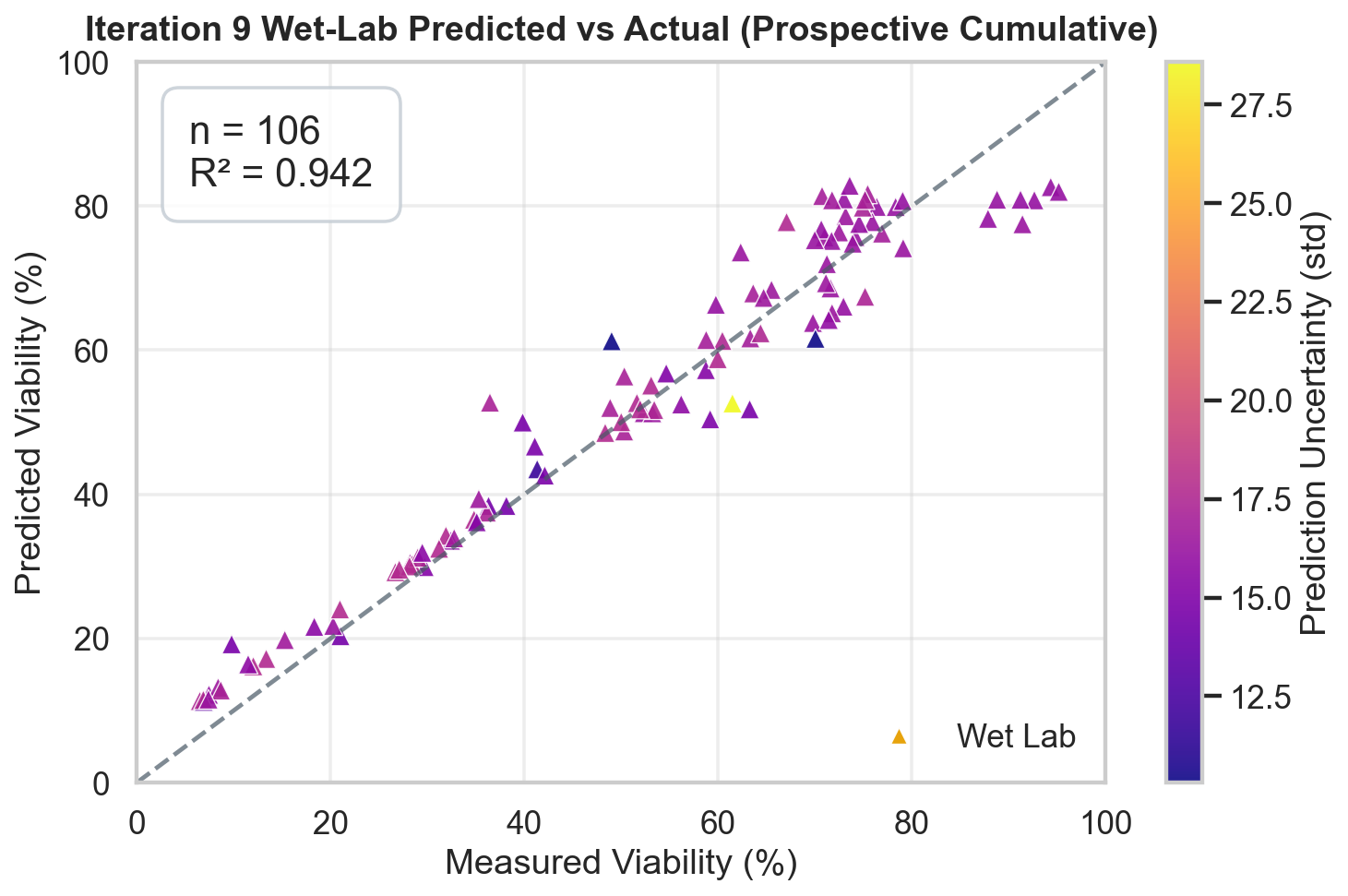}
\end{subfigure}
\hfill
\begin{subfigure}{0.48\textwidth}
\centering
\panellabel{b}
\includegraphics[width=\linewidth]{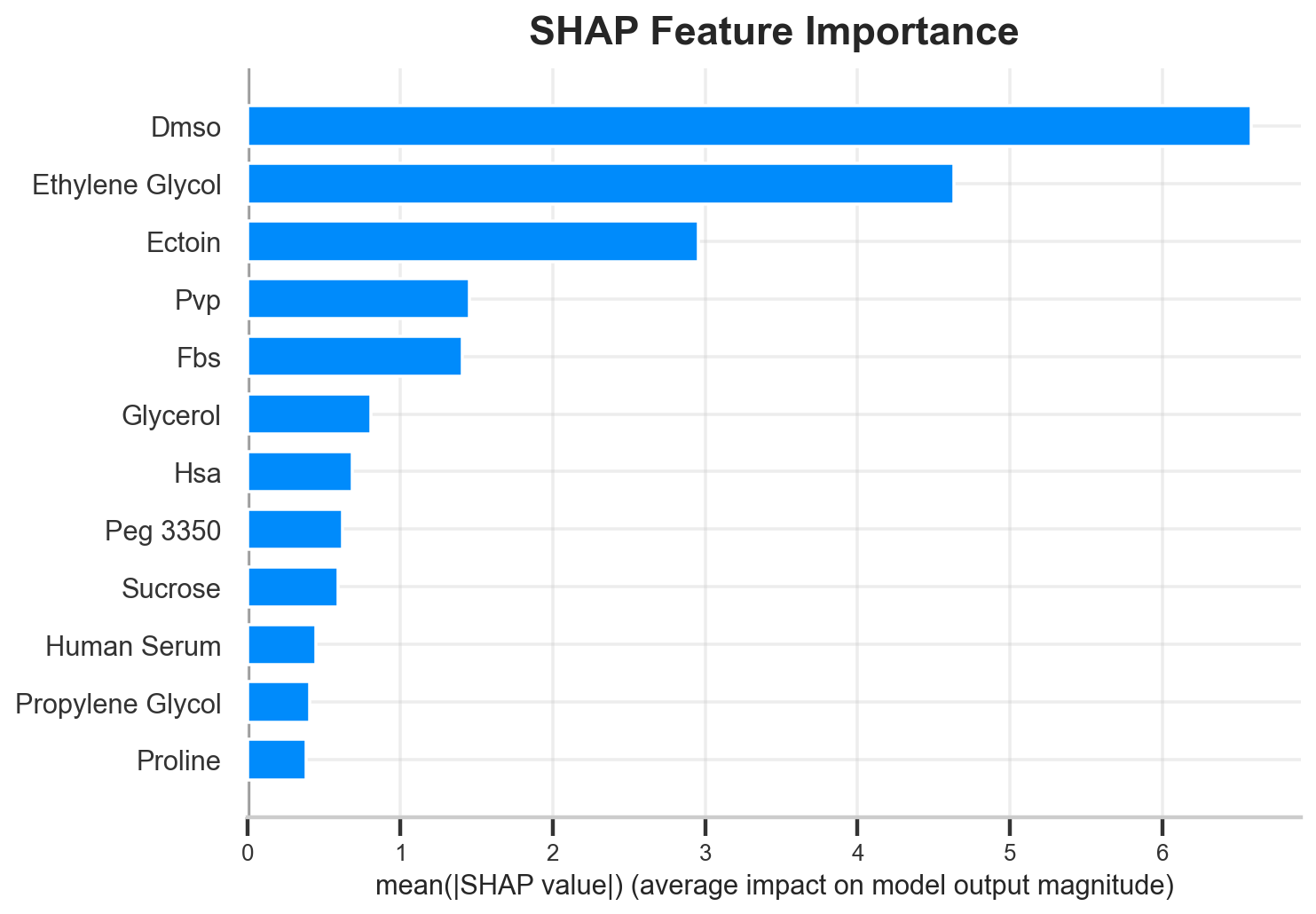}
\end{subfigure}
\caption*{\textbf{Fig. 5 | }Latest wet-lab-corrected model summary after iterative validation: a, cumulative predicted-versus-measured wet-lab viability summary using the iteration-9 checkpoint across 106 wet-lab observations; b, iteration-10 SHAP importance after updating the iteration-9 model with wet-lab validation results. DMSO remained an influential feature, while ethylene glycol and ectoin emerged as major wet-lab-adapted contributors.}
\end{figure}

Apart from regular performance metrics, in each batch, we generated the acquisition landscape and interaction contours that are central to quickly understanding the potentially unexplored areas and ingredient interactions. These plots translate a high-dimensional Gaussian-process model into decision maps that an experimentalist can act on. The acquisition landscape shows where the model predicts high viability, where it is uncertain, and where the combination of the two gives high upper-confidence-bound utility. The script automatically displays the acquisition landscape of the two most important cell-viability determinants. In the latest model's acquisition landscape (\figref[a]{6}), the highest utility region remains concentrated around the ectoin/ethylene glycol neighborhood that had already produced strong wet-lab results, but the uncertainty panel also identifies nearby areas where additional tests would be most informative. This makes the next experimental decision less arbitrary: one can choose candidates trending toward the high-prediction region to confirm reproducibility, or choose candidates near the high-uncertainty/high-UCB boundary to test whether the optimum can be extended.

The interaction contours provide a complementary view. The script automatically selects the most important cell-viability determinant and compares its interactions with the next three most important determinants. In the latest model's interaction contour (\figref[b]{6}), the ethylene glycol/ectoin panel is the most directly actionable because it maps the main chemistry repeatedly selected by the model and validated in the wet lab; the high-response region suggests that future batches should refine this pair locally rather than scan the full formulation space again. The ethylene glycol/DMSO contour is also important because it helps define whether DMSO is acting as a necessary co-protectant or whether low-DMSO formulations can remain viable when ethylene glycol and ectoin are tuned appropriately. Similarly, the ethylene glycol/FBS contour highlights a potentially useful serum-dependent effect, suggesting that FBS may not simply be a background supplement but may modify the concentration range in which ethylene glycol performs well.

Together, these two types of figures support a practical next-step strategy. The acquisition landscape identifies where to test next, while the interaction contours suggest why those tests are biologically and formulation-wise interesting. The most compelling follow-up experiments would therefore not be isolated single recipes, but small local grids around ectoin/ethylene glycol/FBS with controlled low-DMSO levels, plus boundary probes in regions where the model has high uncertainty but plausible acquisition value. Such experiments would directly test whether the apparent optimum is robust, whether DMSO can be further reduced, and whether the serum/ethylene glycol interaction represents a reproducible formulation effect rather than a batch-specific artifact.

\begin{figure}[!htbp]
\centering
\phantomsection\label{fig:6}
\begin{subfigure}{\textwidth}
\centering
\panellabel{a}
\includegraphics[width=\linewidth]{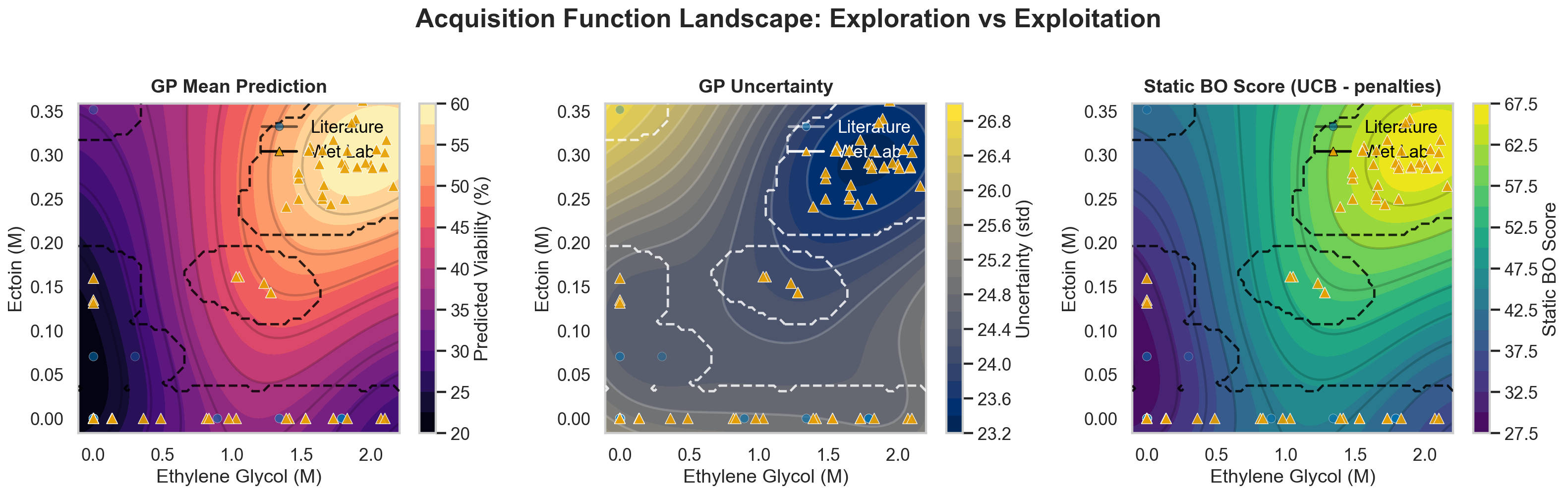}
\end{subfigure}

\vspace{0.8em}

\begin{subfigure}{\textwidth}
\centering
\panellabel{b}
\includegraphics[width=\linewidth]{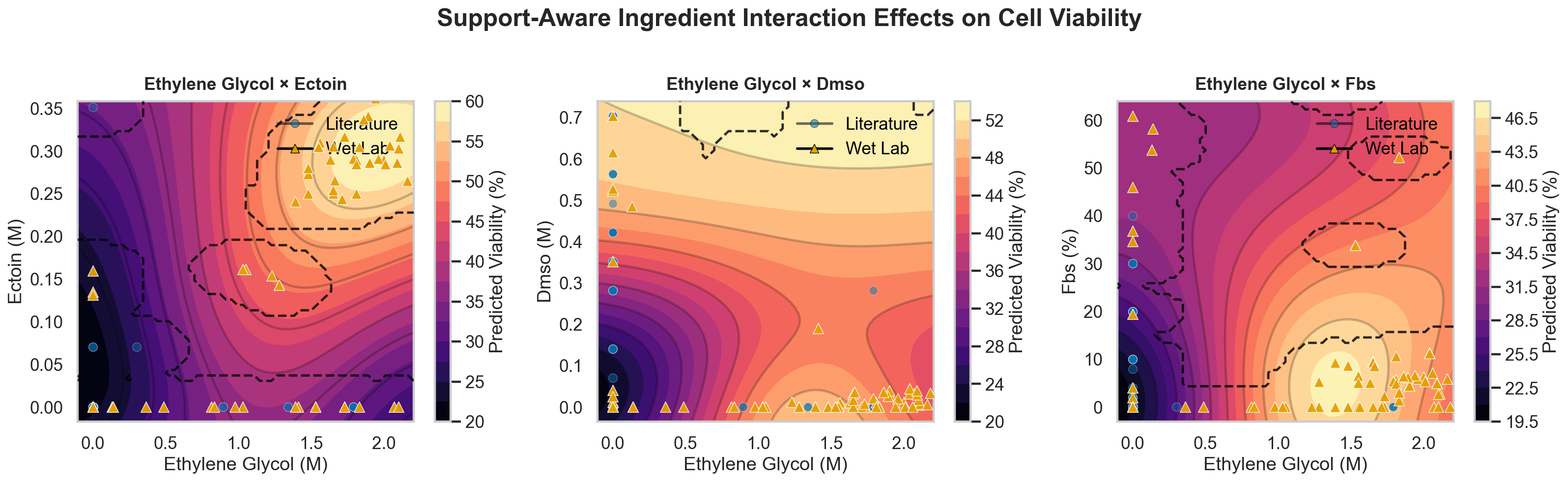}
\end{subfigure}
\caption*{\textbf{Fig. 6 | }Acquisition and interaction landscapes after updating the iteration-9 model: a, predicted mean viability, predictive uncertainty, and static Bayesian-optimization score across the ectoin/ethylene glycol slice. The static Bayesian-optimization score combines UCB with implementation penalties and is therefore not identical to pure UCB (Supplementary Methods); b, support-aware interaction contours for selected ingredient pairs. Filled contour colors indicate model-predicted viability on the displayed two-feature slice while other ingredients are held at the plotting baseline. Dashed boundaries indicate the support-aware region used to distinguish interpolation from extrapolation. Circles denote literature observations and triangles denote wet-lab observations.}
\end{figure}

\FloatBarrier

\section{Discussion}

Owing to the rapid development of large language models (LLMs), this study demonstrates that the lack of in-house computational specialization is no longer an absolute barrier to using machine learning in experimental formulation research. With a single initial prompt, Claude Opus 4.5 generated a runnable baseline repository that translated a scientific objective into data parsing, Gaussian-process modeling, Bayesian optimization with constrained candidate generation, validation templates, and written documentation. Subsequent work with GPT-5.3-codex and human review then converted that first scaffold into a more rigorous closed-loop workflow with iteration-aware model checkpoints, wet-lab-weighted residual correction, explainability plots, and batch recommendation. The important point is that the first answer was useful enough as an executable starting point for a real experimental campaign.\cite{wang_codeact_2024,yang_swe_agent_2024,lu_end_to_end_2026}

This mode of AI assistance is particularly valuable in projects like cryomicroneedle formulation discovery because the computational burden is broad rather than defined by one algorithm. The workflow required many small but connected engineering tasks: synonym handling, unit conversion, duplicate management, model serialization, candidate filtering, uncertainty reporting, wet-lab data ingestion, stage-level evaluation, and figure generation. These are tasks that can consume substantial researcher time even when the underlying scientific question is clear. In our case, the AI-generated infrastructure helped turn a manually curated literature table into a sequential optimization system that ultimately identified simple two- to three-component formulations with high post-thaw viability. These simple formulation results were not what we initially expected given the 10-ingredient limit, but it was a natural optimization outcome; in this small-data setting, high-order combinations are poorly supported and mostly extrapolative, so the model naturally favors sparse formulations where the viability signal is better constrained. In the end, we are happy to see that the assistance provided by LLM agents was not only faster coding, but also a faster conversion of biological intent into a functional and reproducible computational workflow.

Still, the present work is not a demonstration of full scientific autonomy. Literature search and extraction persisted as major human bottlenecks. Search application programming interfaces (APIs) such as Semantic Scholar and OpenAlex lower the cost of finding papers, but they do not guarantee that the relevant corpus is complete, that full text is accessible, or that experimental details are preserved in a machine-readable form.\cite{fricke_semantic_scholar_2018,priem_openalex_2022} Many useful cryopreservation papers reported formulations in tables, scanned PDFs, supplemental files, or inconsistent prose that make rule-based automation error-prone. Literature search and extraction, even with LLM agents, can confuse absolute viability with viability normalized to a control, miss units, merge chemically distinct additives, or treat basal media as active formulation variables. These errors change the training data and therefore change the experiments suggested by the model. For this reason, manual curation and domain-aware checking remained necessary.

Apart from wet-lab validation, which remained fundamentally human and physical, manuscript writing continued to be a human responsibility. LLM agents were helpful for drafting, restructuring, checking code history, and linking computational outputs to figures, but the authors still had to decide what the study could honestly claim. That included distinguishing a proof-of-concept result from a clinic-ready formulation, arguing how useful the results are and how they should be interpreted, as well as describing limitations without overstating the autonomy of the system. In scientific writing, the hardest task may not be producing sentences, but deciding which claims are justified and presentable.

On the experimental front, the most important limitation is that post-thaw viability did not by itself predict cryomicroneedle usability, partly because the present optimization was single-objective. In our validations, most of the formulations that exceeded 70\% cell viability did not form an intact cryomicroneedle patch. A plausible reason is that the formulation chemistry that protects cells during freezing can conflict with the formulation chemistry needed to form a mechanically stable frozen microneedle. High concentrations of permeating cryoprotectants, serum, polymers, and osmolytes can depress the freezing point, increase the unfrozen liquid fraction, alter ice-crystal structure, change viscosity, and/or weaken the frozen matrix. These effects may preserve cells but prevent formation of a rigid, continuous needle array capable of demolding, handling, and inserting into skin. As a result, the high viability we observed should be interpreted as a biological success rather than a complete cryomicroneedle formulation success. The model explicitly optimized post-thaw viability while using DMSO limits and ingredient-count constraints as guardrails, but it did not optimize mechanical strength, patch integrity, and insertion efficiency. This workflow was appropriate for a first proof-of-concept, but it also means that the model could have favored formulations that are biologically protective yet physically unsuitable for cryomicroneedle delivery.\cite{whaley_overview_2021,murray_chemical_2022,chang_cryomicroneedles_2021}

Future work should therefore move from a viability-only formulation search to a multi-objective active-learning workflow. The next model should treat post-thaw viability, intact patch formation, fracture force, and successful skin insertion as coupled outcomes.\cite{shahriari_human_2016,tom_self_driving_2024} Some endpoints could be modeled as continuous objectives, whereas others, such as intact patch formation, may be very useful as a hard feasibility filter. Safety constraints should also be added explicitly, including stricter bounds on ethylene glycol and DMSO, dose-normalized exposure estimates, and searches for chemically safer substitutes that preserve the beneficial ectoin/ethylene-glycol region without carrying the same translational burden.\cite{murray_chemical_2022} We hypothesize that a diversification of ingredients can have a cooperative effect in reducing the concentration of each ingredient, and that a reduced concentration across the board will render forming intact cryomicroneedles tenable.
\newpage
\bibliographystyle{unsrt}
\bibliography{agentic_formulation_discovery}

\end{document}


\section*{Supplementary Methods}

\subsection*{AI-assisted workflow construction}

The initial one-shot prompt and agent transcript were recorded in \texttt{prompt.md} in the GitHub repository. Project code is available at \url{https://github.com/baitmeister/ML-for-CryoMN}. A condensed transcript is provided below with local machine-specific absolute paths replaced by relative project-root references. The transcript is included only to document the first computational scaffold; final workflow components already described in the main manuscript are not repeated here.

\begingroup
\small
\setlength{\parskip}{0.45em}
\begin{promptbar}
\subsubsection*{Condensed chat conversation}

Note: \textit{Long scientific background already covered in the main manuscript and routine file-edit confirmations are omitted except where they document a change in workflow logic.}

\paragraph*{User Input}

The scientific background, optimization targets and major constraints from the initial prompt are summarized in the main manuscript rather than repeated here. Implementation-specific prompt details retained for reproducibility were that the agent should parse formulation and viability text into a structured matrix, merge supplied synonym groups, exclude culture media as variables, harmonize concentration units where possible, handle duplicate formulations with discordant viability values, and output scripts plus Markdown README files for each major step. The complete original wording is preserved in \texttt{prompt.md}.

\textit{Listed directory [ML for CryoMN](project-root/)}

\paragraph*{Planner Response}

Found the CSV file. Let me examine its contents to understand the data structure and then develop an appropriate ML approach.

\textit{Viewed [Cryopreservative Data 2026.csv](project-root/Cryopreservative\%20Data\%202026.csv)}

\paragraph*{Planner Response}

Let me see the rest of the file to understand the full dataset and then create a comprehensive plan.

\textit{Viewed [Cryopreservative Data 2026.csv](project-root/Cryopreservative\%20Data\%202026.csv)}

\paragraph*{Planner Response}

Now I have a comprehensive understanding of the data. Let me create the implementation plan detailing the ML approach and the first step (data parsing). The dataset has \textasciitilde{}900 rows which is relatively small, so I need to choose an appropriate ML approach.

\textit{Edited relevant file}

\textit{Edited relevant file}

\paragraph*{Planner Response}

I've created the implementation plan. Now I need to notify the user for their review and approval before proceeding to execution.

\paragraph*{User Input}

This is a system-generated message that is sent due to the user's review policy. Proceed with the decision that you think is the most optimal here.

\textit{User accepted the command \texttt{mkdir -p "ML for CryoMN/data/raw" "ML for CryoMN/data/processed" "ML for CryoMN/data/validation" "ML for CryoMN/src/01\_data\_parsing" "ML for CryoMN/src/02\_model\_training" "ML for CryoMN/src/03\_optimization" "ML for CryoMN/src/04\_validation\_loop" "ML for CryoMN/docs"}}

\textit{User accepted the command \texttt{cp "ML for CryoMN/Cryopreservative Data 2026.csv" "ML for CryoMN/data/raw/"}}

\paragraph*{Planner Response}

Now let me create the main parsing script. This is a comprehensive script that will parse the formulation data from the CSV file.

\textit{Edited relevant file}

\paragraph*{Planner Response}

Now let me create the README documentation for the parsing step and also create a requirements.txt file:

\textit{Edited relevant file}

\textit{Edited relevant file}

\textit{Edited relevant file}

\paragraph*{Planner Response}

Now let me run the parsing script to test it and verify it works correctly:

\textit{User accepted the command \texttt{cd "ML for CryoMN" \&\& python src/01\_data\_parsing/parse\_formulations.py 2>\&1 | head -100}}

\paragraph*{Planner Response}

The script found 169 duplicate pairs which is too many to resolve interactively. I need to modify the script to handle this better - by automatically using the mean viability for duplicates or keeping the first occurrence. Let me update the script:

\paragraph*{Planner Response}

I need to update the parsing script to handle duplicates automatically instead of requiring interactive input for 169 pairs. Let me fix this:

\textit{Viewed [parse\_formulations.py](project-root/src/01\_data\_parsing/parse\_formulations.py)}

\paragraph*{Planner Response}

Now I need to update the script to handle duplicates automatically. I'll modify the duplicate resolution function to automatically average viabilities for duplicate formulations instead of requiring interactive input:

\textit{Edited relevant file}

\paragraph*{Planner Response}

I need to use the proper format for replace\_file\_content with the TargetContent parameter:

\textit{Edited relevant file}

\paragraph*{Planner Response}

Now I need to update the main function to call the new [auto\_resolve\_duplicates](cci:1://project-root/src/01\_data\_parsing/parse\_formulations.py:489:0-552:13) function instead of the old interactive one:

\textit{Edited relevant file}

\paragraph*{Planner Response}

Now let me run the parsing script again to verify it works correctly:

\textit{User accepted the command \texttt{cd "ML for CryoMN" \&\& python src/01\_data\_parsing/parse\_formulations.py}}

\paragraph*{Planner Response}

The script ran successfully! It parsed 374 formulations, found 169 duplicate pairs, averaged them down to 200 unique formulations, and saved the output. However, I notice a viability value of 341\% which seems invalid (likely a relative viability compared to control). Let me verify the output and check on this issue:

\textit{User accepted the command \texttt{cd "ML for CryoMN" \&\& head -20 data/processed/parsed\_formulations.csv}}

\textit{User accepted the command \texttt{cd "ML for CryoMN" \&\& python -c "import pandas as pd; df = pd.read\_csv('data/processed/parsed\_formulations.csv'); print('Rows with viability > 100\%:'); print(df[df['viability\_percent'] > 100][['formulation\_id', 'viability\_percent', 'source\_doi']].head(10))"}}

\paragraph*{Planner Response}

I see there are some entries with viability > 100\% (like 341\% which is an average including values "130\% compared to standard DMSO"). These are relative viabilities compared to a control, not absolute percentages. I need to handle this edge case. Let me also update the task checklist and create a walkthrough document:

\textit{Edited relevant file}

\textit{Edited relevant file}

\paragraph*{Planner Response}

I've completed Step 1: Data Parsing. Let me notify the user about the completed work and share the results:

\paragraph*{User Input}

Proceed, do not ask me for permission in the future. Complete the entire workflow

\textit{Edited relevant file}

\textit{Edited relevant file}

\paragraph*{Planner Response}

Now let me run the model training script:

\textit{User accepted the command \texttt{cd "ML for CryoMN" \&\& python src/02\_model\_training/train\_gp\_model.py}}

\paragraph*{Planner Response}

Model training completed successfully! The GP model achieved:
\begin{itemize}
\item CV R\textsuperscript{2}: 0.239 (indicating the task is challenging with this limited data)
\item Training R\textsuperscript{2}: 0.686
\item DMSO is the most important feature
\end{itemize}

Now let me proceed with Step 3: Optimization:

\textit{Edited relevant file}

\textit{Edited relevant file}

\paragraph*{Planner Response}

Now let me run the optimization script:

\textit{User accepted the command \texttt{cd "ML for CryoMN" \&\& python src/03\_optimization/optimize\_formulation.py}}

\paragraph*{Planner Response}

The optimization is running in background. Let me check its status:

\textit{Checked command status}

\paragraph*{Planner Response}

Still running. Let me wait a bit more:

\textit{Checked command status}

\paragraph*{Planner Response}

There's an error - the candidates DataFrame is empty because the optimization isn't finding valid candidates. Let me fix the optimize function:

\textit{Viewed [optimize\_formulation.py](project-root/src/03\_optimization/optimize\_formulation.py)}

\paragraph*{Planner Response}

I see the issue - the \texttt{differential\_evolution} might not be converging properly. Let me fix the optimize function to handle empty candidates and simplify the optimization approach:

\textit{Edited relevant file}

\paragraph*{Planner Response}

Now let me run the optimization again:

\textit{User accepted the command \texttt{cd "ML for CryoMN" \&\& python src/03\_optimization/optimize\_formulation.py}}

\paragraph*{Planner Response}

Excellent! Optimization completed successfully! Now let me create Step 4: Validation Loop and finish the workflow:

\textit{Edited relevant file}

\textit{Edited relevant file}

\paragraph*{Planner Response}

Now let me run the validation loop script to create the template, then update the task checklist and walkthrough:

\textit{User accepted the command \texttt{cd "ML for CryoMN" \&\& python src/04\_validation\_loop/update\_model.py}}

\paragraph*{Planner Response}

Now let me update the task.md and walkthrough.md with the final results:

\textit{Edited relevant file}

\textit{Edited relevant file}

\paragraph*{Planner Response}

Now let me update the main project README with complete documentation and verify everything is working correctly:

\textit{Edited relevant file}

\textit{User accepted the command \texttt{cd "ML for CryoMN" \&\& ls -la results/}}

\textit{User accepted the command \texttt{cd "ML for CryoMN" \&\& cat results/candidates\_general\_summary.txt}}

\paragraph*{Planner Response}

Excellent! The entire pipeline is complete and working. Let me view the DMSO-free candidates and then notify the user with the final summary:

\textit{User accepted the command \texttt{cd "ML for CryoMN" \&\& cat results/candidates\_dmso\_free\_summary.txt}}

\paragraph*{User Input}

have you run the whole pipeline once? if not please run it to see if there's any error

\paragraph*{Planner Response}

Yes, I ran each step individually during development! But let me run the complete pipeline from scratch end-to-end to verify there are no errors:

\textit{User accepted the command \texttt{cd "ML for CryoMN" \&\& echo "=== Step 1: Data Parsing ===" \&\& python src/01\_data\_parsing/parse\_formulations.py \&\& echo "" \&\& echo "=== Step 2: Model Training ===" \&\& python src/02\_model\_training/train\_gp\_model.py \&\& echo "" \&\& echo "=== Step 3: Optimization ===" \&\& python src/03\_optimization/optimize\_formulation.py \&\& echo "" \&\& echo "=== Step 4: Validation Loop ===" \&\& python src/04\_validation\_loop/update\_model.py \&\& echo "" \&\& echo "=== PIPELINE COMPLETE ==="}}

\paragraph*{Planner Response}

\textbf{\(\checkmark\) Complete pipeline ran successfully with no errors!}

All 4 steps executed without any issues:

\begin{center}
\begin{tabular}{lll}
\hline
Step & Status & Key Output \\ \hline
\textbf{1. Data Parsing} & \(\checkmark\) & 200 unique formulations extracted \\
\textbf{2. Model Training} & \(\checkmark\) & GP model trained (CV R\textsuperscript{2}=0.24, Training R\textsuperscript{2}=0.69) \\
\textbf{3. Optimization} & \(\checkmark\) & 20 general + 15 DMSO-free candidates generated \\
\textbf{4. Validation Loop} & \(\checkmark\) & Template created, awaiting wet lab results \\
\hline
\end{tabular}
\end{center}

\textbf{Key Results:}
\begin{itemize}
\item \textbf{Best general candidate}: 78.6\% \(\pm\) 21.1\% viability at 0.5\% DMSO
\item \textbf{Best DMSO-free candidate}: 77.9\% \(\pm\) 23.5\% viability at 0\% DMSO
\end{itemize}

The pipeline is ready for use. Step 4 correctly indicates it's waiting for wet lab validation results before it can update the model.

\end{promptbar}
\endgroup

\subsection*{Dataset parsing and feature encoding}

The first step was to identify data points with ‘unique conditions’. These included: (i) experiments conducted with the same formulations tested on different cell types, (ii) experiments where pre-treatment was done, or specialised methods such as electroporation was used to introduce the cryoprotectant into the cell, (iii) experiments conducted with the same polymer or copolymer of different molecular weights/ratios, (iv) studies where viability or recovery metrics was reported as a value relative to a standard experiment, and (v) experiments where commercial cryoprotectants was used whose full formations is unknown or not reported. This  manual process reduced usable data points
from 499 to 393 (Supplementary Data).

Free-text formulation records were parsed into canonical ingredient features. Synonyms were merged for common cryoprotectants and additives, including DMSO/Me\(_2\)SO, ethylene glycol/EG, propylene glycol/1,2-propanediol/PROH, ectoin/ectoine, FBS/FCS, HES/hydroxyethyl starch, HSA/human serum albumin, methylcellulose/MC, hyaluronic acid/HMW-HA and dextran/dextran-40. Basal media and buffers, including DMEM, \(\alpha\)-MEM, PBS, HBSS, saline and related media terms, were excluded as formulation variables.

Molar-convertible ingredients were represented in molar units (\texttt{\_M}). Percentage-only ingredients, including sera, proteins, PEG species, polymers and other complex mixtures, were represented as percentage features (\texttt{\_pct}). PEG was split by molecular weight class rather than treated as one generic feature. Concentration parsing supported percentage, M, mM and \(\mathrm{mg}/\mathrm{ml}\) inputs where molecular weights and densities permitted conversion.

Rows without extractable formulation text or extractable viability were excluded from model training. When identical canonical formulation signatures were associated with different extracted viabilities, the duplicate signatures were collapsed by averaging the viability values and retaining one representative formulation row. The final parsed literature dataset contained 198 unique formulations and 34 canonical formulation features. For active model training, features were retained only if present in at least three literature formulations, yielding 21 active input variables.

\subsection*{Gaussian-process model configuration}

The literature-only surrogate used Gaussian-process regression with a Mat\'ern kernel with smoothness parameter \(\nu=2.5\), where \(\nu\) controls the smoothness of the fitted response function. The kernel also included a constant-amplitude term and a white-noise term. Input features were standardized using \texttt{StandardScaler}, the target was normalized internally by the Gaussian-process regressor and hyperparameter optimization used 10 optimizer restarts. The initial model used five-fold cross-validation with \texttt{shuffle=True} and \texttt{random\_state=42}.

\subsection*{Prior-mean correction update}

After wet-lab data became available, the production update strategy used a composite model. A literature Gaussian process first predicted a baseline viability for each formulation. A second Gaussian process was then fitted to wet-lab residuals, where residual means measured viability minus literature-model prediction. The composite prediction was:

\begin{equation}
\hat{y}(x)=\hat{y}_{\mathrm{literature}}(x)+\hat{r}_{\mathrm{wetlab}}(x),
\end{equation}

where \(x\) is the formulation vector, \(\hat{y}(x)\) is final predicted viability, \(\hat{y}_{\mathrm{literature}}(x)\) is the literature-model prediction and \(\hat{r}_{\mathrm{wetlab}}(x)\) is the wet-lab residual correction. The combined uncertainty was computed by summing the literature and correction variances and taking the square root.

Source-specific noise assumptions were fixed at \(\alpha_{\mathrm{literature}}=1.0\) and \(\alpha_{\mathrm{wetlab}}=0.02\), where \(\alpha\) is the Gaussian-process observation-noise regularization parameter. This corresponds to a 50-fold relative weighting in favour of wet-lab observations. Wet-lab generalization was estimated by cross-validating the wet-lab rows only, with \(K=\min(5,n_{\mathrm{wetlab}})\), where \(K\) is the number of folds and \(n_{\mathrm{wetlab}}\) is the number of available wet-lab rows. All literature rows were retained in the training set for every fold. Cross-validated residuals were used to estimate an additive prediction-bias correction (\texttt{bias\_shift\_percent}) and a multiplicative uncertainty correction (\texttt{uncertainty\_scale}), which were applied consistently during optimization, evaluation and next-batch recommendation.

\subsection*{Random-sampling candidate generation}

In the first five stages, a large random pool of formulations was sampled from the feasible formulation space. Candidates were normalized, trace concentrations were removed and hard constraints were applied before prediction. Candidates were then ranked by Gaussian-process predicted mean viability. We used two predefined DMSO-capped candidate pools: a general pool allowing up to 5\% v/v DMSO, and a project-defined “DMSO-free” pool allowing up to 0.5\% v/v DMSO. In this manuscript, “DMSO-free pool” refers to this operational low-DMSO optimization stratum rather than strict chemical absence. Candidate identity used a practical concentration floor: percentage features below 0.1\% and molar features below 0.001 M, equivalent to 1 mM, were treated as absent.

\subsection*{Acquisition-score scale and prediction calibration}

For Bayesian optimization, the acquisition score was computed on the original post-thaw viability scale, not on a standardized target scale. Feature standardization was applied internally before Gaussian-process prediction, but both the predictive mean and uncertainty used for acquisition remained in viability percentage-point units.

For literature-only checkpoints, the uncalibrated prediction was the mean and standard deviation returned by the literature-trained Gaussian process. For wet-lab-updated checkpoints, the prior-mean correction model described above supplied the uncalibrated composite mean \(\mu_{0}(x)\) and standard deviation \(\sigma_{0}(x)\). After each wet-lab update, cross-validated wet-lab residuals were used to estimate an additive bias correction \(b\) and a multiplicative uncertainty scale \(s\), with \(s\) selected to improve empirical one- and two-standard-deviation coverage.

The calibrated prediction used for acquisition scoring was therefore
\begin{equation}
\mu(x)=\mu_{0}(x)+b,
\end{equation}
\begin{equation}
\sigma(x)=s\,\sigma_{0}(x).
\end{equation}
If a checkpoint did not contain calibration metadata, the defaults were
\begin{equation}
b=0,
\qquad
s=1.
\end{equation}
For the iteration-10 checkpoint, for example, the stored calibration parameters were
\begin{equation}
b=-6.83 \text{ percentage points},
\qquad
s=0.78.
\end{equation}

The upper-confidence-bound (UCB) acquisition score was then computed as
\begin{equation}
a_{\mathrm{UCB}}(x)
=
\mu(x)+\kappa\sigma(x),
\end{equation}
with \(\kappa=0.5\). Predictions were not clipped to the interval $0$--$100\%$ before acquisition scoring. Exported candidate tables reported the calibrated predicted viability, calibrated uncertainty, and acquisition value for each candidate.

\subsection*{Bayesian optimization candidate generation}

Bayesian optimization used the same Gaussian-process or composite surrogate model but replaced random candidate search with Differential-Evolution optimization of the calibrated upper-confidence-bound score defined above. Each run generated 20 candidates with \texttt{maxiter=100}, \texttt{popsize=15} and \texttt{random\_seed=42}.

The objective also included implementation penalties to keep candidates chemically plausible and experimentally useful: a sparsity penalty of 0.35 for formulations with more active ingredients than the observed reference count, a support penalty of 4.0 for candidates outside the observed formulation manifold, and a local batch-diversity penalty of 5.0 with radius 0.05 to discourage near-duplicate candidates within the same slate. The support radius was the weighted 90th percentile of observed nearest-neighbour distances multiplied by 1.25. Search bounds were ingredient-specific: DMSO was bounded by the selected cap, ethylene glycol/glycerol/propylene glycol by 2.5 M, trehalose/sucrose/raffinose by 1.0 M, proline/betaine/ectoin-class osmolytes by 0.5 M, serum features by 90\%, and hyaluronic acid/methylcellulose by 2\%.

\subsection*{Reasoning for optimization method}

We used a staged active-learning strategy that transitioned from broad constrained random candidate generation toward Bayesian-optimization-guided candidate generation as wet-lab evidence accumulated. The initial phase used random sampling to explore the experimentally feasible formulation space under practical constraints. Bayesian-optimization candidate files were generated during the transition period and were audited against wet-lab testing outcomes, but the early prospective batches were not treated as purely acquisition-driven optimization campaigns.

This design was recommended by Claude and chosen because the initial model was trained largely from literature-derived formulations, which provided useful prior structure but was not yet sufficiently calibrated to the specific wet-lab assay, formulation constraints, and experimental handling conditions used in this study. In this setting, applying Bayesian optimization too aggressively at the outset would risk over-exploiting artefacts of the literature prior, inaccurate uncertainty estimates, or sparsely supported regions of the high-dimensional formulation space.

The random-search phase therefore served as a burn-in period. Candidate formulations were sampled broadly under practical constraints and then evaluated prospectively, producing wet-lab observations that were not concentrated around a single predicted optimum. This broad sampling improved coverage of the experimentally accessible formulation space, exposed systematic model residuals, and provided the wet-lab anchors needed to recalibrate prediction bias and uncertainty. It also reduced the risk that the optimization trajectory would collapse prematurely onto a narrow region supported only by the literature model.

After this initial wet-lab grounding phase, the accumulated experimental data provided a more reliable basis for model-guided optimization. Later stages increasingly used Differential-Evolution-based Bayesian optimization together with the dedicated next-batch generator. The Bayesian-optimization module searched for high-acquisition candidate pools, while the next-batch generator converted those pools into wet-lab slates with an adaptive exploitation/exploration split, residual-based blind-spot probes, local-rank probes and coverage probes.

This two-phase strategy separates model grounding from model exploitation. Broad random sampling was used while uncertainty estimates and rankings were still being established; acquisition-guided search was introduced once the model had sufficient prospective wet-lab evidence to make model-guided selection scientifically meaningful. The approach therefore prioritizes robustness in the early phase and sample efficiency in the later phase.

\subsection*{Cryomicroneedle mold specification}

The PDMS mold was made using a 1:9 ratio of curing agent and base supplied in the SYLGARD 184 Silicone Elastomer kit. Following vacuuming to remove air bubbles, the PDMS with the positive mold was cured in an oven at \(60\,^{\circ}\mathrm{C}\) for 60 minutes. The positive mold was 3-D printed with the following specification: a \(10\times10\) pyramidal microneedle array patch; patch dimensions of \(1.63\,\mathrm{cm}\times1.63\,\mathrm{cm}\) with \(0.25\,\mathrm{cm}\) margins on each side; pitch \(=1200\,\mu\mathrm{m}\); needle base \(=500\,\mu\mathrm{m}\); needle height \(=1600\,\mu\mathrm{m}\); volume per needle \(=(1/3)\times500\times500\times1600\,\mu\mathrm{m}^{3}\approx0.133\,\mu\mathrm{L}\); and volume per patch \(=100\times0.133\approx13.3\,\mu\mathrm{L}\).

\subsection*{Experimental batch design}

The next-batch generator treated Bayesian optimization as a candidate-pool source rather than as the complete experimental policy. It always wrote a 20-formulation slate. The default split was 8 exploitation rows and 12 exploration rows, with exploitation count adaptively bounded between 4 and 12 according to the previous completed stage's RMSE, mean signed residual and empirical \(1\sigma\) coverage. Exploitation candidates were selected from the active-stage Bayesian-optimization candidate files after removing already tested formulations and applying a chemistry-family diversity cap.

Exploration candidates came from local-rank probes, blind-spot probes and coverage probes. Local-rank probes perturbed dominant ingredients around high-value anchors. Blind-spot probes targeted historically underpredicted regions using positive residual thresholds evaluated in descending order: 10, 8, 5, 2 and 0 percentage points. Exactly two coverage probes were reserved by default and selected by greedy \(k\)-center distance from the observed context. The generator also wrote recommended subsets for wet-lab capacities from 6 to 12 formulations.

\subsection*{Stage-based evaluation}

Stage evaluation used frozen model checkpoints. Each stage was evaluated against the wet-lab batch selected by that stage's own candidate outputs. For example, the literature-only model was evaluated against the first wet-lab batch selected from literature-only candidate files, while iteration 3 was evaluated against the formulations selected from iteration 3 outputs. The evaluation did not use later models to rescore earlier decisions as if those later models had been available at the time.

For each stage batch, the evaluation script reported root mean squared error (RMSE), mean absolute error (MAE), Spearman's rank correlation coefficient \(\rho\), Kendall's rank correlation coefficient \(\tau\), mean predictive standard deviation, empirical coverage within \(\pm 1\sigma\) and \(\pm 2\sigma\), and threshold-decision hit rates at 50\% and 70\% viability. Note that only the first four were reported in the main text.

Specifically, model predictions were evaluated on the wet-lab-tested formulations, where \(y_i\) is the measured post-thaw viability, and viability and uncertainty were expressed in percentage-point units:
\begin{equation}
\mathcal{D}_{s}=\{(x_i,y_i)\}_{i=1}^{n_s}
\end{equation}

Then prediction error was quantified using root mean squared error (RMSE) and mean absolute error (MAE), where \(\mu_i\) is the model-predicted mean viability:
\begin{equation}
\mathrm{RMSE}_{s}=\sqrt{\frac{1}{n_s}\sum_{i=1}^{n_s}(y_i-\mu_i)^2}
\end{equation}
\begin{equation}
\mathrm{MAE}_{s}=\frac{1}{n_s}\sum_{i=1}^{n_s}|y_i-\mu_i|
\end{equation}
Lower RMSE and MAE indicate more accurate viability prediction. RMSE penalizes large errors more strongly because residuals are squared, whereas MAE reports the average absolute prediction error in the original viability scale.

Ranking performance was assessed using Spearman's rank correlation coefficient and Kendall's rank correlation coefficient. Spearman's \(\rho\) was computed as the Pearson correlation between the ranks of measured and predicted viability:
\begin{equation}
\rho_s =
\frac{
\sum_{i=1}^{n_s}
\left(R(y_i)-\overline{R(y)}\right)
\left(R(\mu_i)-\overline{R(\mu)}\right)
}{
\sqrt{
\sum_{i=1}^{n_s}\left(R(y_i)-\overline{R(y)}\right)^2
}
\sqrt{
\sum_{i=1}^{n_s}\left(R(\mu_i)-\overline{R(\mu)}\right)^2
}
}
\end{equation}
Higher \(\rho_s\) indicates better preservation of the ordering of formulations by viability; \(\rho_s=1\) indicates perfect monotonic agreement, \(\rho_s=0\) indicates no monotonic association, and negative values indicate inverse ranking.

Kendall's \(\tau\) was computed from pairwise ordering agreement:
\begin{equation}
\tau_s =
\frac{C_s-D_s}
{\sqrt{(C_s+D_s+T_y)(C_s+D_s+T_\mu)}}
\end{equation}
where \(C_s\) and \(D_s\) are the numbers of concordant and discordant formulation pairs, respectively, and \(T_y\) and \(T_\mu\) denote ties only in measured and predicted viability. Higher \(\tau_s\) indicates better pairwise ranking consistency.

Predictive uncertainty was summarized by the mean predicted standard deviation:
\begin{equation}
\overline{\sigma}_{s}=\frac{1}{n_s}\sum_{i=1}^{n_s}\sigma_i
\end{equation}
This metric reports the average model uncertainty for the evaluated formulations. It is diagnostic rather than directly optimized: lower values indicate narrower predictive distributions, but uncertainty should be interpreted jointly with empirical coverage and error.

Calibration was assessed by empirical one-standard-deviation coverage:
\begin{equation}
\mathrm{Coverage@1\sigma}_{s}
=
\frac{1}{n_s}
\sum_{i=1}^{n_s}
\mathbf{1}\left(|y_i-\mu_i|\leq\sigma_i\right)
\end{equation}
For well-calibrated Gaussian predictive intervals, this value is expected to be approximately \(0.68\). Values below \(0.68\) indicate overconfident uncertainty estimates or systematic prediction bias, whereas values above \(0.68\) indicate conservative uncertainty estimates.

Threshold decision performance was evaluated at viability thresholds of \(50\%\) and \(70\%\). For a threshold \(\theta\), the hit rate was defined as:
\begin{equation}
\mathrm{HitRate}_{s}(\theta)
=
\frac{1}{n_s}
\sum_{i=1}^{n_s}
\mathbf{1}
\left[
\mathbf{1}(\mu_i\geq\theta)
=
\mathbf{1}(y_i\geq\theta)
\right],
\qquad
\theta\in\{50,70\}
\end{equation}

Candidate files and next-formulation slates were cross-referenced against wet-lab rows by canonical formulation signatures after applying the same trace-concentration floor used during candidate generation.

\subsection*{Prospective-cumulative $R^2$}

For predicted-versus-actual model-performance reporting, $R^2$, the coefficient of determination, was computed using a prospective-cumulative protocol. At stage \(k\), where \(k\) is the current model stage, predictions from the frozen stage model were evaluated against the cumulative wet-lab rows available through stages \(0 \ldots k\). This preserves temporal ordering while increasing sample size and outcome-range coverage. Batch-specific $R^2$ was not used as a primary endpoint because individual wet-lab batches were small and often had restricted response ranges, making $R^2$ sensitive to outliers and sign changes that did not necessarily reflect practical predictive utility. The stage-indexed predicted-versus-actual plots used measured viability on the x-axis, predicted mean viability on the y-axis and predictive standard deviation as point colour.

\subsection*{Explainability analysis}

Explainability analyses were generated from the active iteration-specific observed context, combining literature and wet-lab rows with source weights. Feature importance was computed by weighted permutation importance using degradation in weighted $R^2$. SHAP analysis used \texttt{KernelExplainer} when the SHAP package was available, with up to 100 background samples and up to 100 explained samples. The feature-importance overview and SHAP plots were limited to the top 12 features, empirical marginal plots to the top 6 features and interaction contours to the top 3 feature pairs.

Support-aware plots (on Github) used weighted observed quantile ranges from the 1st to 95th percentile rather than raw extrema. In one-dimensional empirical marginal plots, stronger local support was indicated by solid line segments and weaker support by dashed segments. In two-dimensional interaction and acquisition plots, dashed boundaries indicated stronger pairwise empirical support. The static acquisition landscape reused the Bayesian-optimization scoring convention, including UCB with \(\kappa=0.5\), support penalty 4.0, support-radius scale 1.25 and sparsity penalty 0.35, but excluded sequential batch-diversity penalties because those depend on already selected candidates. The uncertainty dashboard summarized predicted-versus-actual agreement, empirical-versus-nominal coverage over uncertainty multipliers from 0.5 to 2.5, absolute error versus predictive standard deviation and mean uncertainty across viability bands.

\subsection*{Figure and artifact traceability}

The computational figures referenced in this supplementary analysis were generated directly from GitHub repository artifacts. Stage-level evaluation outputs were written under \texttt{results/evaluation/}, including \texttt{stage\_performance.png}, \texttt{next\_formulations\_performance.png}, \texttt{single\_objective\_progress.png} and \texttt{iteration\_prospective\_metrics.csv}. Stage-indexed predicted-versus-actual $R^2$ plots were written under \texttt{results/explainability/stage\_r2/}. Iteration-specific explainability outputs were written under \texttt{explainability/<iteration\_tag>/}, including \texttt{feature\_importance.png}, \texttt{shap\_summary.png}, \texttt{shap\_importance.png}, \newline \texttt{partial\_dependence\_plots.png}, \texttt{interaction\_contours.png}, \texttt{acquisition\_landscape.png}, \texttt{uncertainty\_analysis.png} and \texttt{support\_diagnostics.png}.